\documentclass[10pt,journal]{IEEEtran}

\usepackage[utf8]{inputenc}
\usepackage[T1]{fontenc}
\usepackage{microtype}

\usepackage[table,dvipsnames]{xcolor}

\usepackage{amsmath}
\usepackage{amssymb}
\usepackage{mathtools}
\usepackage{amsthm}
\usepackage{amsfonts}
\usepackage{nicefrac}
\usepackage{bm}

\usepackage{booktabs}       
\usepackage{multirow}
\usepackage{threeparttable}
\usepackage{colortbl}
\usepackage{array}
\usepackage{tabularx}
\newlength{\krslotwidth}
\newlength{\crslotwidth}
\newcommand{\setkrcrwidths}[2]{%
  \settowidth{\krslotwidth}{\textbf{#1}}%
  \settowidth{\crslotwidth}{\textbf{#2}}%
}
\newcommand{\krcr}[2]{(\makebox[\krslotwidth][r]{#1},\hspace{0.15em}\makebox[\crslotwidth][r]{#2})}

\usepackage{graphicx}
\usepackage{wrapfig}
\usepackage{stfloats}

\makeatletter
\@ifundefined{algorithm}{}{}
\@ifundefined{endalgorithm}{}{}

\expandafter\ifx\csname algorithm*\endcsname\relax\else
  \expandafter\let\csname algorithm*\endcsname\relax
\fi
\expandafter\ifx\csname endalgorithm*\endcsname\relax\else
  \expandafter\let\csname endalgorithm*\endcsname\relax
\fi

\@ifundefined{listofalgorithms}{}{}

\@ifundefined{algorithmic}{}{}
\makeatother

\usepackage[linesnumbered,ruled,vlined]{algorithm2e}
\SetKwComment{Comment}{$\triangleright$\ }{}
\SetKwRepeat{Do}{do}{while}

\SetKwInOut{Input}{Input}
\SetKwInOut{Output}{Output}
\SetNlSkip{0.5em}
\usepackage{url}
\usepackage{tcolorbox}
\usepackage{enumitem}
\usepackage{balance}
\usepackage{textcomp}
\usepackage{verbatim}
\usepackage{cite}

\definecolor{LinkBlue}{RGB}{25, 88, 166}
\definecolor{LinkPurple}{RGB}{150, 0, 80}
\usepackage[colorlinks,linkcolor=LinkPurple,citecolor=LinkBlue,urlcolor=black]{hyperref}

\usepackage[capitalize,noabbrev]{cleveref}

\theoremstyle{plain}
\newtheorem{theorem}{Theorem}[section]

\newtheorem{corollary}[theorem]{Corollary}
\theoremstyle{definition}

\theoremstyle{remark}

\usepackage[disable,textsize=tiny]{todonotes}

\renewcommand{\paragraph}[1]{\smallskip\noindent\textbf{#1.}}
\providecommand{\subparagraph}[1]{}
\renewcommand{\subparagraph}[1]{\noindent\textbf{\underline{#1}}}

\newif\ifshowzhnotes
\showzhnotesfalse
\ifshowzhnotes
  \usepackage{CJKutf8}
  \AtBeginDocument{\begin{CJK*}{UTF8}{gbsn}}
  \AtEndDocument{\clearpage\end{CJK*}}
  \long\def\zh#1{#1}
\else
  \long\def\zh#1{}
\fi

\newcommand{\Wq}{\bm{W}_\mathrm{q}}
\newcommand{\Wk}{\bm{W}_\mathrm{k}}
\newcommand{\Wv}{\bm{W}_\mathrm{v}}
\newcommand{\Wo}{\bm{W}_\mathrm{o}}

\newcommand{\mrm}[1]{\mathrm{#1}}
\newcommand{\mbb}[1]{\mathbb{#1}}

\begin{document}

\title{ResidualKV: Residual-Based KV Cache Compression \\ for Efficient Long-Context Inference}

\author{
  Jitai Hao, Qiang Huang, Yaowei Wang, Min Zhang, Jun Yu 
  \thanks{This work was supported by the National Natural Science Foundation of China (NSFC) under Grant Nos. 62125201, U24B20174, and U25B6003, as well as the New Generation Artificial Intelligence-National Science and Technology Major Project (2025ZD0123302). (Corresponding authors: Qiang Huang, Jun Yu)}
  \thanks{Jitai Hao, Qiang Huang, and Jun Yu are with the School of Intelligence Science and Engineering, Harbin Institute of Technology (Shenzhen), Shenzhen, China. (email: jitaihao@outlook.com, \{huangqiang, yujun\}@hit.edu.cn)}
  \thanks{Yaowei Wang and Min Zhang are with the School of Computer Science and Technology, Harbin Institute of Technology (Shenzhen), Shenzhen, China. (email: \{wangyaowei, zhangmin2021\}@hit.edu.cn)}
}

\markboth{Preprint}%
{Hao \MakeLowercase{\textit{et al.}}: ResidualKV: Residual-Based KV Cache Compression for Efficient Long-Context Inference}


\maketitle 

\begin{abstract}
Efficient long-context inference faces two coupled bottlenecks: KV-cache memory grows linearly with context length, while attention computation grows quadratically.
Existing approaches typically address one at the expense of irreversible token eviction, full-cache retention, or full-history reconstruction, limiting their effectiveness for multi-turn interaction and long-form reasoning.
Motivated by two empirical properties, \emph{Long-Range} Inter-Token Similarity and Smooth Residual Distribution, we propose \textbf{ResidualKV}, which factorizes the KV cache into a sparse set of globally retrieved references and compact, quantized residual codes for the remaining tokens.
This representation preserves token-specific information without permanent eviction and, when combined with sparse attention, reconstructs only the selected states on demand. Dynamic-stride scheduling further reduces reference growth from linear to approximately logarithmic at ultra-long contexts.
Across Llama, Qwen, LLaVA-OV, and Qwen3-VL backbones, ResidualKV maintains near-full-cache performance using only \textbf{13\%--16\%} KV storage and \textbf{30\%} attention computation on LongBench, and \textbf{8\%--10\%} storage and \textbf{10\%} computation in matched-budget multimodal evaluation. It also accelerates decoding by up to \textbf{1.5}$\times$ with KV-cache quantization and \textbf{3.4}$\times$ without it.
These results show that global cross-token redundancy supports accurate, memory-efficient, and computation-efficient long-context inference.
The source code is available at \url{https://github.com/CURRENTF/ResidualKV}.
\end{abstract}

\begin{IEEEkeywords}
KV Cache Compression, Long-Context Inference, Residual Compression, Efficient Inference.
\end{IEEEkeywords}

\section{Introduction}
\label{sec:intro}

Long-context foundation models, including large language models (LLMs) and multimodal large language models (MLLMs), increasingly underpin persistent intelligent assistants and autonomous agents for long-document and code analysis, scientific discovery and deep research, multimodal understanding, and embodied decision-making \cite{yao2023react, jimenez2024swe, shi2025deep, zhu2026deeprubric, alayrac2022flamingo, guha2023legalbench, hao2026unix, wu2024longvideobench, fu2025videomme, tian2024drivevlm}.
These applications involve three complementary long-horizon workloads: long-input understanding over extensive documents, code, or visual observations; multi-turn interactions with evolving information needs over a shared context; and long-output reasoning with continually accumulating historical states.

\begin{figure}[t]
  \centering
  \includegraphics[width=0.99\linewidth]{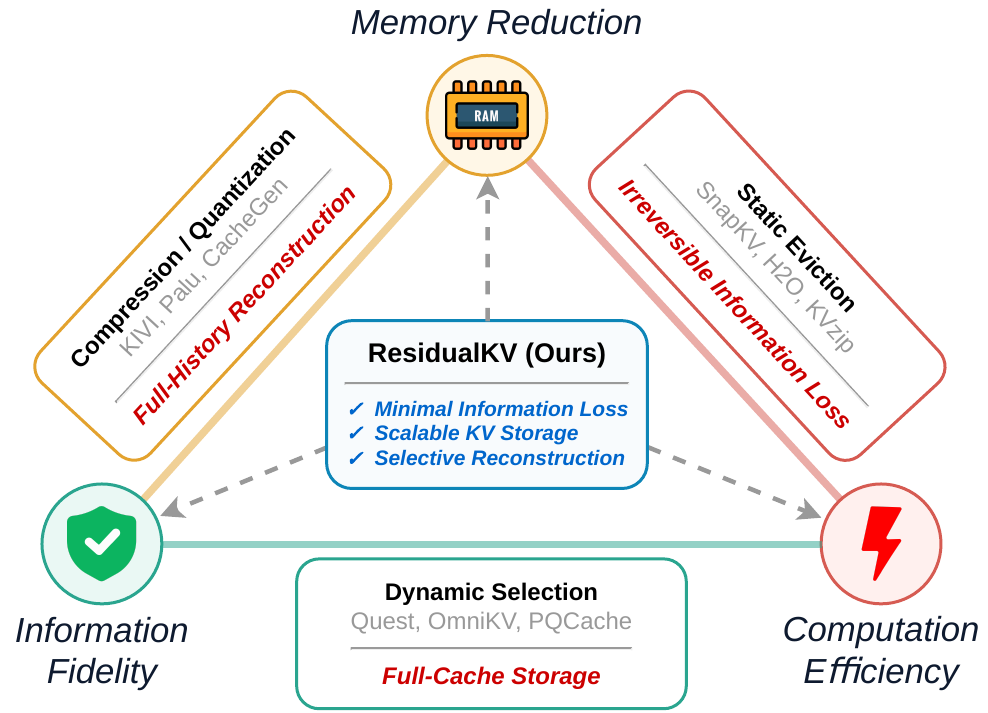}
  \caption{\textbf{Three-way trade-off in KV cache optimization.} 
  Existing method families typically balance only two of information fidelity, memory reduction, and computational efficiency, whereas ResidualKV jointly achieves all three.}
  \label{fig:tradeoff_radar}
  \vspace{-1.0em}
\end{figure}

Efficiently serving long-context workloads is challenging: attention computation grows quadratically with sequence length, while KV-cache storage scales linearly. 
For example, the KV cache of Qwen2.5-32B~\cite{team2024qwen2} exceeds \textbf{270 GB} at a context length of 128K and a batch size of 8, far beyond a single GPU's capacity.
Consequently, inference systems must jointly preserve historical information, minimize KV-cache storage, and reduce attention computation. 
Existing approaches \cite{li2024snapkv, zhang2023h2o, kim2025kvzip, tang2024quest, hao2025omnikv, zhang2025pqcache, liu2024kivi, chang2025palu, liu2024cachegen, oren2024transformers_tova, liu2023scissorhands, liu2025clusterkv}, however, typically optimize only a subset of these goals, resulting in a three-way trade-off (Figure~\ref{fig:tradeoff_radar}):
\begin{itemize}[left=0pt, topsep=0.25em, itemsep=0.15em]
  \item \textbf{Inreversible Information Loss:}
  Static eviction \cite{li2024snapkv, zhang2023h2o} permanently discards seemingly unimportant tokens. Since future queries are unpredictable, this causes irreversible performance degradation in evolving workloads like multi-turn dialogue~\cite{li2025scbench} and long-chain reasoning~\cite{maa_aime}.

  \item \textbf{Heavy Memory Storage:}
  Dynamic KV selection \cite{tang2024quest, hao2025omnikv, zhang2025pqcache} reduces compute by sparsely attending to query-relevant tokens. Yet, it still requires the complete history to remain accessible, leaving memory capacity or offloading bandwidth as a primary bottleneck.

  \item \textbf{Low Computation Efficiency:}
  Quantization \cite{liu2024kivi, zandieh2025turboquant, hooper2024kvquant} and representation compression \cite{chang2025palu, liu2024cachegen, kim2025lexico} reduce storage but typically require full-history reconstruction. This leaves attention computation largely unoptimized and introduces substantial recovery overhead.
\end{itemize}

Jointly achieving near-lossless performance, compact KV storage, and efficient computation remains an open challenge.

\begin{figure}[t]
  \centering
  \includegraphics[width=0.99\linewidth]{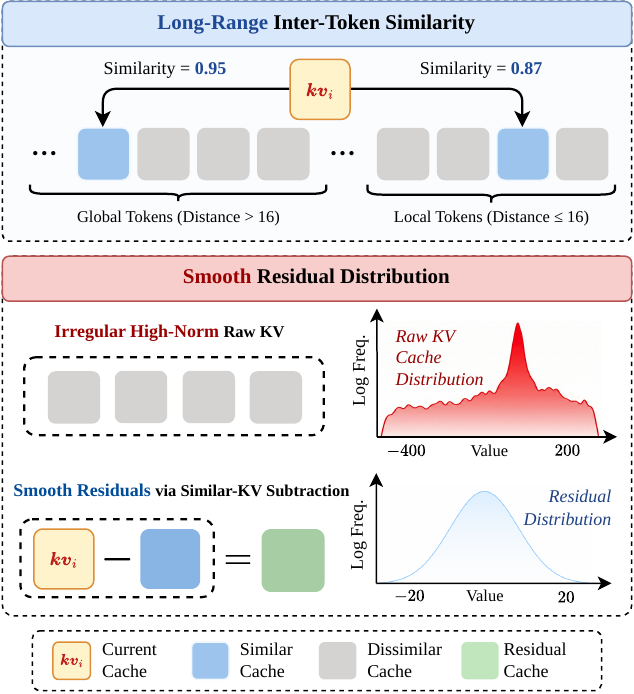}
  \caption{\textbf{Two empirical properties underlying ResidualKV.} 
  KV states exhibit \emph{long-range inter-token similarity}, with highly similar references often located beyond local neighborhoods. Subtracting these references removes shared components and yields a \emph{smooth residual distribution} concentrated near zero, facilitating compression and low-bit quantization.}
  \label{fig:teaser}
  \vspace{-1.5em}
\end{figure}

To address this challenge, we revisit cross-token redundancy in KV representations and uncover two empirical properties, illustrated in Figure~\ref{fig:teaser}.
\textbf{(1) \emph{Long-Range} Inter-Token Similarity:} 
Existing compression and merging methods primarily exploit local redundancy through adjacent-token encoding, contiguous merging, or neighborhood-based KV aggregation~\cite{liu2024cachegen, wang2024kvmerger, hu2025efficient_chelsea, cui2025homogeneous, yuan2025weightedkv}. 
In contrast, we find that highly similar KV states frequently occur at distant contextual positions, making historical tokens effective global references.
\textbf{(2) \emph{Smooth} Residual Distribution:} 
Subtracting a retrieved reference removes the component shared across similar KV states. 
The remaining token-specific residual has a substantially smaller magnitude, concentrates near zero, and contains fewer extreme values than the original state, making it more amenable to dimensionality reduction and low-bit quantization.
These observations motivate a reference-residual factorization of the KV cache: reusable cross-token structure can be retained in a sparse set of global references, while the information unique to each token can be stored as a compact, quantization-friendly residual.

Based on these observations, we propose \textbf{ResidualKV}, a residual-based KV cache compression framework compatible with both low-bit quantization and sparse attention.
ResidualKV retains a small set of uncompressed reference tokens and represents each remaining token by a compact residual obtained by subtracting the mean of several globally similar references.
Its asymmetric heavy-compressor, lightweight-decompressor design shifts computation to one-time compression while minimizing repeated reconstruction costs during decoding.
Fixed-stride scheduling provides a near-lossless default, while dynamic-stride scheduling progressively increases reference spacing to limit memory growth in ultra-long contexts.
Combined with selection methods such as OmniKV \cite{hao2025omnikv} and SnapKV \cite{li2024snapkv}, ResidualKV reconstructs only the tokens required by the current attention operation, avoiding full-history reconstruction, memory transfer, and computation.

The main contributions are summarized as follows:
\begin{itemize}[left=0pt, topsep=0.25em, itemsep=0.15em]
  \item \textbf{Representation-Level Insights:}
  We reveal pervasive long-range similarity among KV states and show that subtracting globally retrieved references produces smooth, low-magnitude residuals. 
  These findings expose a previously underexplored form of global cross-token redundancy that is well suited to compression and quantization.

  \item \textbf{Unified ResidualKV Framework:}
  We develop a general reference-residual representation that integrates global reference retrieval, compact residual encoding, low-bit quantization, scalable reference scheduling, and selective on-demand reconstruction. 
  By preserving both shared structure and token-specific deviations, ResidualKV reduces storage without permanently discarding historical information.

  \item \textbf{Efficient and Flexible Inference:}
  ResidualKV is compatible with static and dynamic sparse-attention methods and reconstructs only query-selected states using a lightweight decompressor. It thereby jointly reduces attention computation, unnecessary decompression, and KV cache memory traffic.
  The compressor and decompressor add fewer than \textbf{5\%} parameters and require about \textbf{8 GPU-hours} of training.
\end{itemize}

We systematically evaluate ResidualKV across long-input understanding, multi-turn interaction, long-output reasoning, and multimodal understanding. 
Across Llama, Qwen, LLaVA-OV, and Qwen3-VL backbones, ResidualKV maintains near-full-cache performance with only \textbf{13\%--16\%} KV storage and \textbf{30\%} attention computation on LongBench, and \textbf{8\%--10\%} storage and \textbf{10\%} computation in matched-budget multimodal settings.
It further improves decoding throughput by up to \textbf{1.5}$\times$ over full attention.
Overall, ResidualKV unifies information preservation, memory compression, and sparse computation for efficient long-context inference.

The remainder of this paper is organized as follows. 
Section~\ref{sec:related_work} reviews related work. 
Section~\ref{sec:observations} presents the empirical observations underlying ResidualKV. Section~\ref{sec:method} describes the ResidualKV framework. 
Section~\ref{sec:expt} reports experimental results, and finally, Section~\ref{sec:conclusions} concludes the paper.

\section{Related Work}
\label{sec:related_work}

As foundation models scale, extensive research seeks to improve inference efficiency and support long-horizon reasoning and interaction~\cite{li2024snapkv, oren2024transformers_tova, liu2024intactkv, lin2025quantization, qi2025deltallm, liu2024deepseekv2, guo2025enhancing, hao2025tokenLRC, yan2025inftythink, yan2026inftythinkplus, lu2026uicopilot, NEURIPS2025_02d028c3}.
The directions most relevant to ResidualKV include Token Selection and Eviction, Multimodal Token Reduction, KV Cache Compression, Token Clustering, and KV Cache Quantization, each balancing information fidelity, memory reduction, and computational efficiency differently.

\paragraph{Token Selection and Eviction}
Token selection and eviction reduce attention computation by processing only a subset of historical tokens.
Static methods \cite{zhang2023h2o, oren2024transformers_tova, li2024snapkv, xiao2024efficient, kim2025kvzip} permanently evict tokens using importance, reconstruction, or early-attention signals. 
For example, SnapKV~\cite{li2024snapkv} retains high-attention tokens, whereas KVzip~\cite{kim2025kvzip} identifies query-agnostic tokens through context reconstruction with paraphrased prompts.
However, discarded information cannot be recovered for future queries, limiting these methods in multi-turn dialogue and long-chain reasoning.
Dynamic methods \cite{hao2025omnikv, tang2024quest, zhang2025pqcache, xiao2024infllm, liu2025deepseek32, sun2025shadowkv} instead select tokens according to each decoding query.
Quest~\cite{tang2024quest} selects KV pages using element-wise extrema, while OmniKV~\cite{hao2025omnikv} exploits inter-layer attention similarity to guide sparse attention using a few observation layers.
Although more robust to changing information needs, these methods generally retain the full historical cache, often via CPU offloading, and thus remain constrained by storage and PCIe transfer.
ResidualKV complements both families by compressing rather than discarding historical states and reconstructing only the selected tokens on demand.

\paragraph{Multimodal Token Reduction}
MLLMs exhibit substantial redundancy among image and video tokens, motivating visual-sequence reduction.
FastV~\cite{chen2024image_fastv} prunes visual tokens after early layers based on concentrated visual attention; PACT~\cite{dhouib2025pact} combines pruning with clustering; DivPrune~\cite{alvar2025divprune} prioritizes visually diverse evidence; and FastVID~\cite{shen2025fastvid} dynamically adjusts token density during video inference.
These methods shorten visual inputs but do not compress the KV representations accumulated during long-context or autoregressive decoding.
ResidualKV instead operates on the KV cache representation itself and applies to both textual and multimodal tokens, making it orthogonal and complementary to visual-token reduction.

\paragraph{KV Cache Compression}
KV cache compression exploits low-rank, codebook, or reconstruction structures to reduce storage \cite{saxena2024eigen, zhang2024lorc, chang2025palu, liu2024deepseekv2, kim2025kvzip, kim2025lexico}. 
Palu~\cite{chang2025palu} projects KV states into a low-rank latent space, LoRC~\cite{zhang2024lorc} retains principal components obtained through SVD, and Lexico~\cite{kim2025lexico} represents KV states through sparse coding over learned dictionaries.
However, these methods typically reconstruct or process the entire compressed history during decoding, introducing repeated projection or decoding overhead.
ResidualKV represents each token using global references and a compact token-specific residual; combined with sparse attention, it reconstructs only query-selected tokens ($\leq$10\% in our sparse multimodal configurations) while maintaining near-full-cache performance.

\paragraph{Token Clustering} 
Token clustering exploits inter-token redundancy by reusing or aggregating similar KV states.
ClusterKV~\cite{liu2025clusterkv} relies on CPU-resident clustering structures and incurs PCIe traffic, whereas Chelsea~\cite{hu2025efficient_chelsea} restricts matching to local neighborhoods and may overlook distant but related states.
ResidualKV instead retrieves global, long-range references, subtracts their shared components, and stores a compact residual for each token, preserving token-specific information rather than replacing it with a cluster representative.
This residual representation is composable with static eviction, dynamic selection, multimodal token reduction, low-rank projection, and quantization, as evaluated in Section~\ref{sec:expt:analysis}.

\paragraph{KV Cache Quantization}
KV cache quantization reduces memory footprint and data movement by representing cached states at lower numerical precision \cite{liu2024kivi, xiao2023smoothquant, hooper2024kvquant, he2024zipcache}. 
KIVI~\cite{liu2024kivi} applies asymmetric, channel-wise key quantization to mitigate outliers, while KVQuant~\cite{hooper2024kvquant} quantizes pre-RoPE keys using learned non-uniform codebooks.
However, channel- or block-wise layouts can hinder fine-grained sparse access by requiring unselected data to be loaded and dequantized.
ResidualKV produces low-magnitude, near-zero residuals suitable for token-wise low-bit quantization, further reducing storage while supporting selective, on-demand reconstruction.

\section{Observations}
\label{sec:observations}

\begin{figure*}[t]
\centering
\begin{minipage}[b]{0.32\textwidth}
  \centering
  \includegraphics[width=\linewidth]{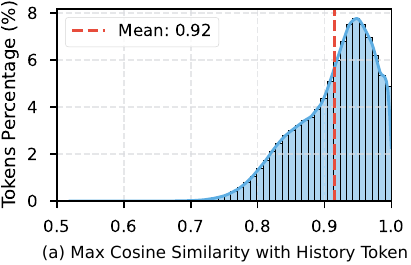} 
\end{minipage}
\hfill
\begin{minipage}[b]{0.32\textwidth} 
  \centering
  \includegraphics[width=\linewidth]{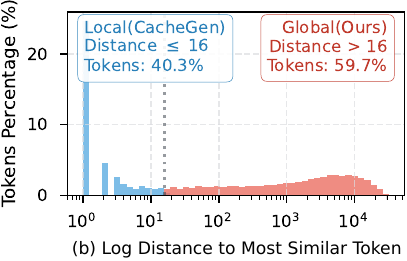}
\end{minipage}
\hfill
\begin{minipage}[b]{0.32\textwidth}
  \centering
  \includegraphics[width=\linewidth]{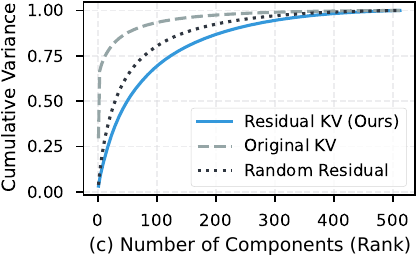}
\end{minipage}
\begin{minipage}[b]{0.32\textwidth}
  \centering
  \includegraphics[width=\linewidth]{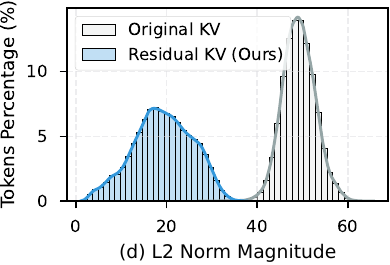}
\end{minipage}
\hfill
\begin{minipage}[b]{0.32\textwidth}
  \centering
  \includegraphics[width=\linewidth]{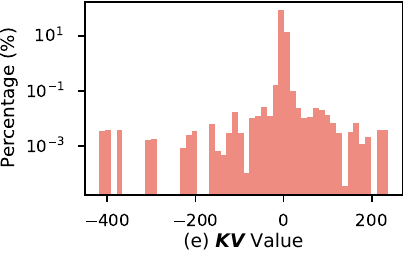}
\end{minipage}
\hfill
\begin{minipage}[b]{0.32\textwidth}
  \centering
  \includegraphics[width=\linewidth]{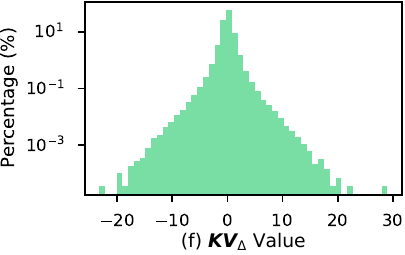}
\end{minipage}
\caption{\textbf{Empirical analysis of cross-token KV redundancy and residualization.}
(a) The histogram of maximum historical cosine similarity for each token. 
(b) The histogram of log distance to the most similar historical token for each token.
(c) Cumulative variance of retrieved-reference residuals, original KV states, and random residuals. 
(d) The histogram of $L_2$ norm distributions of original and residual KV states. 
(e) and (f) Value distributions before and after reference subtraction, respectively; Residualization removes dominant shared components and produces low-magnitude, near-zero token-specific residuals.}
\label{fig:combined_analysis_results}
\vspace{-0.5em}
\end{figure*}

\paragraph{Background and Notation}
ResidualKV targets standard Transformer architectures, including Qwen~\cite{yang2025qwen2_1m} and Llama~\cite{grattafiori2024llama3}, where each layer consists of self-attention and a feed-forward network (FFN).
Let $\Wq \in \mbb{R}^{d \times d_\mrm{q}}$, $\Wk \in \mbb{R}^{d \times d_\mrm{k}}$, $\Wv \in \mbb{R}^{d \times d_\mrm{v}}$, and $\Wo \in \mbb{R}^{d_\mrm{q} \times d}$ denote the query, key, value, and output projections, where $d$ is the hidden dimension and $d_\mrm{q}, d_\mrm{k}, d_\mrm{v}$ are the corresponding dimensions across all heads; typically, $d_\mathrm{k} = d_\mathrm{v}$.
Given token representations $\bm{H} \in \mathbb{R}^{B \times L \times d}$, the key and value states are: 
\begin{displaymath}
  \bm{K}=\bm{H}\Wk~\text{and}~\bm{V}=\bm{H}\Wv,
\end{displaymath}
with shapes $[B, L, d_\mathrm{k}]$ and $[B, L, d_\mathrm{v}]$, respectively. 
When $d_\mathrm{k} = d_\mathrm{v} = ND$, both can be reshaped as $[B, L, N, D]$, where $B$, $L$, $N$, and $D$ denote the batch size, sequence length, number of key/value heads, and head dimension, respectively.

Transformer inference comprises a \emph{prefill} phase followed by autoregressive \emph{decoding}. 
Prefill processes the input sequence and stores its key-value states. 
At decode step $t$, the new query attends to the cached history ($\bm{K}_{<t}$, $\bm{V}_{<t}$), after which the newly generated states ($\bm{K}_{t}$, $\bm{V}_{t}$) are appended to the cache. 
This cache eliminates repeated projection of the entire prefix, but its memory footprint grows linearly with $B$, $L$, and $d_\mathrm{k}$, making it a primary bottleneck in long-context inference.

To characterize cache redundancy, we concatenate the key and value states and define their \textbf{residual representation} as:
\begin{displaymath}
  \bm{KV}_\Delta = \bm{KV} - \overline{\bm{KV}}_\text{ref},
\end{displaymath}
where $\bm{KV} = \mathrm{Concat}(\bm{K}, \bm{V})$ and $\overline{\bm{KV}}_\text{ref}$ is the mean of the $k$ nearest historical KV states selected as references. 
Unless otherwise specified, the following analyses use Qwen2.5-7B-Instruct-1M \cite{yang2025qwen2_1m}, with results reported in Figure \ref{fig:combined_analysis_results}.

\paragraph{Observation 1: \emph{Long-Range} Inter-Token Similarity}
Cross-token redundancy is often treated as local because adjacent textual or visual tokens describe related content, motivating methods like CacheGen~\cite{liu2024cachegen} to encode KV differences within local groups.
However, our analysis reveals strong long-range similarity: a token's maximum cosine similarity to historical states averages 0.92 (Figure~\ref{fig:combined_analysis_results}(a)), and 59.7\% of the nearest matches lie over 16 positions away (Figure~\ref{fig:combined_analysis_results}(b)).
Therefore, effective residual compression should retrieve references \emph{globally} from a compact set to exploit distant redundancy while limiting search and storage overhead.

\paragraph{Observation 2: \emph{Smooth} Residual Distribution}
Beyond identifying similar references, we analyze the representations remaining after their shared components are removed.
Original KV states concentrate variance in dominant high-energy directions encoding shared linguistic, semantic, and structural patterns, yielding rapid cumulative spectral growth (Figure~\ref{fig:combined_analysis_results}(c)).
Residual states instead approach the flatter spectrum of random residuals, showing that reference subtraction removes these shared directions.
Residualization also reduces representational scale and irregularity: residual states have markedly smaller $L_2$ norms (Figure~\ref{fig:combined_analysis_results}(d)), while the large magnitudes and long tail of raw KV states (Figure~\ref{fig:combined_analysis_results}(e)) become a near-zero distribution with far fewer extremes (Figure~\ref{fig:combined_analysis_results}(f)).
The resulting low-energy, token-specific deviations from reusable historical references are thus better suited to compact encoding and low-bit quantization.

\paragraph{Implications for KV-Cache Compression}
These observations motivate a reference-residual factorization of the KV cache: a sparse set of globally distributed reference tokens preserves shared cross-token structure, while each remaining token is represented by a compact residual relative to its retrieved references. 
The first observation motivates \emph{global reference retrieval} and scalable reference scheduling; the second motivates \emph{residual compression} and token-wise low-bit quantization. 
Because each residual remains associated with its original token, the corresponding KV state can also be reconstructed selectively when requested by sparse attention. 
These principles directly lead to the ResidualKV framework introduced next.

\section{The ResidualKV Framework}
\label{sec:method}

\begin{figure*}[t]
  \centering
  \includegraphics[width=0.99\textwidth]{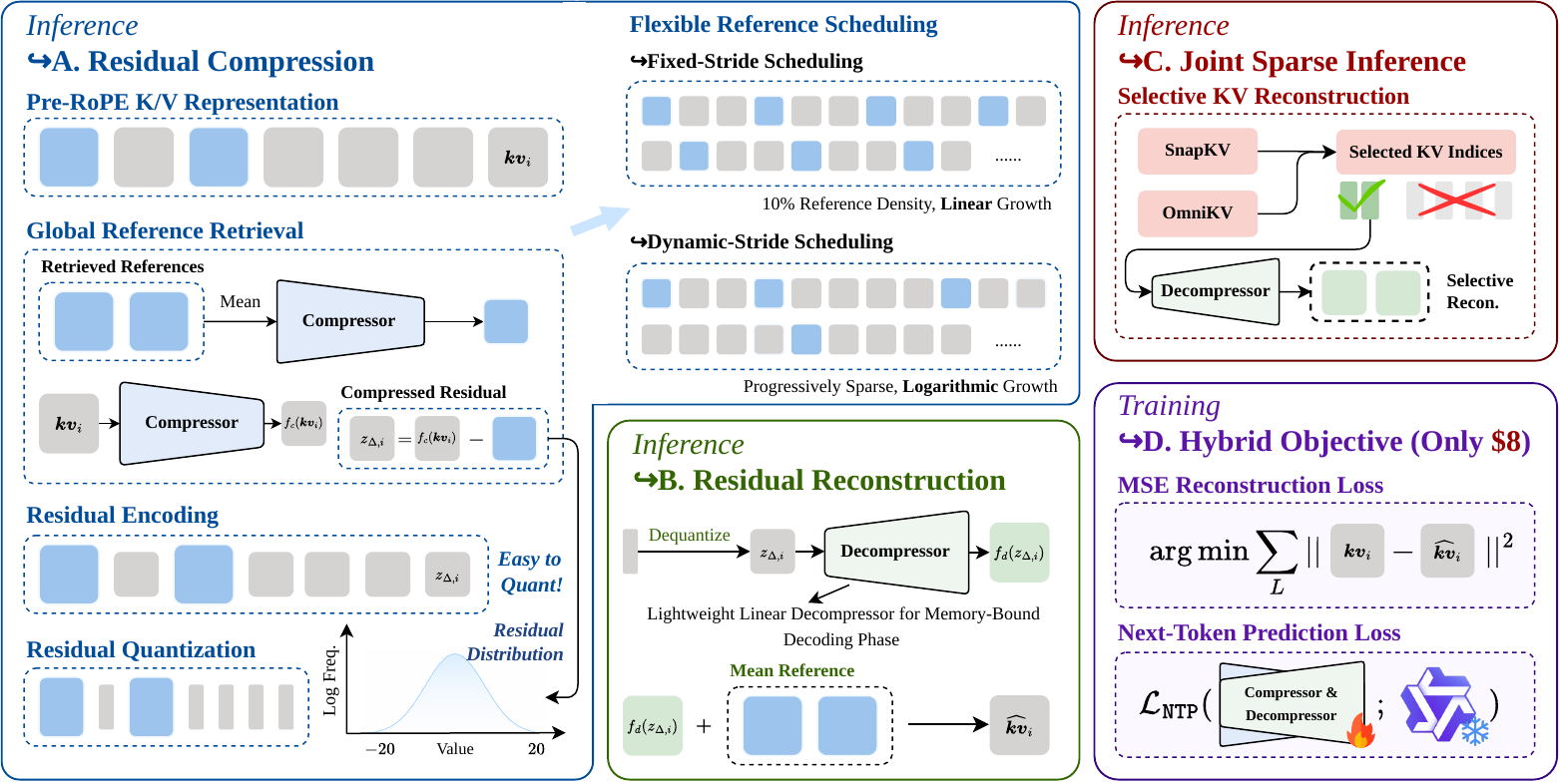}
  \caption{\textbf{Overview of ResidualKV.}
  This framework encompasses both inference and training workflows across four key components: (A) \emph{Residual Compression} (\S\ref{sec:method:compression}), which retrieves reference KV cache via flexible scheduling to generate highly quantizable compressed residuals; (B) \emph{Residual Reconstruction} (\S\ref{sec:method:reconstruction}), which employs a lightweight linear decompressor during the memory-bound decoding phase to reconstruct original KV cache; (C) \emph{Joint Sparse Inference} (\S\ref{sec:method:sparse_inference}), which seamlessly integrates with sparse attention methods to perform selective KV reconstruction; and (D) \emph{Hybrid Objective Training} (\S\ref{sec:method:training}), a low-cost training scheme (costing only \$8) using a joint objective.}
  \label{fig:ResidualKV}
  \vspace{-0.5em}
\end{figure*}

We present \textbf{ResidualKV}, a residual-based KV-cache compression framework illustrated in Figure~\ref{fig:ResidualKV}. 
Its central idea is to separate reusable cross-token structure, captured by globally retrieved references, from token-specific information, encoded as compact residuals. This reference-residual representation avoids directly compressing entangled raw KV states and enables memory-efficient storage with selective reconstruction. 
We next detail its design.

\subsection{Residual Compression}
\label{sec:method:compression}

Residual compression proceeds in four stages: constructing position-invariant K/V representations, retrieving global historical references, encoding token-specific residuals, and optionally applying low-bit quantization.

\paragraph{Pre-RoPE K/V Representation}
ResidualKV operates on pre-RoPE key-value states~\cite{su2024roformer}, preventing position-dependent rotations that distort $L_2$ distances between semantically similar tokens at different positions. 
For token $i$, let $\bm{kv}_i \in \mathbb{R}^{2d_{\text{k}}}$ be the concatenated key and value states across all heads, where typically $d_{\text{k}}=d_{\text{v}}=ND$. 
Because both project the same hidden state, ResidualKV retrieves and compresses them jointly to exploit their shared structure. 
The matched-budget ablation in Figure~\ref{fig:kv_concat_ntp_loss} confirms that joint K/V compression yields lower training NTP loss than separate compression.

\paragraph{Global Reference Retrieval}
ResidualKV retrieves historical references from a compact set $\mathcal{T}_{\text{ref}}$, avoiding a search over the complete token history while retaining access to long-range redundancy. 
For token $i$, we define: 
\begin{displaymath}
  k_i=\min(k,|\mathcal{T}_{\text{ref}}|),
\end{displaymath}
so that retrieval remains well-defined during warm-up before $k$ references are available. 
The top-$k_i$ references are selected based on the squared $L_2$ distance:
\begin{displaymath}
  \mathcal{R}_i = \underset{\bm{kv}_j \in \mathcal{T}_\text{ref}}{\operatorname{arg\,top-}k_i} (-\|\bm{kv}_i - \bm{kv}_j\|_2^2).
\end{displaymath}
Their arithmetic mean defines the representative reference:
\begin{displaymath}
  \overline{\bm{KV}}_{\text{ref},i} = \frac{1}{k_i} \sum_{j\in \mathcal{R}_i} \bm{kv}_j \in \mathbb{R}^{2d_{\text{k}}}.
\end{displaymath}
After warm-up, $k_i=k$, recovering the configured top-$k$ mean. 
Averaging multiple references suppresses token-specific noise, syntactic variation, and activation spikes that may affect a single nearest neighbor. 
Moreover, using $L_2$ distance aligns reference retrieval with the Mean Squared Error (MSE) reconstruction objective, yielding a stable estimate of the shared component.

\paragraph{Residual Encoding}
Given $\overline{\bm{KV}}_{\text{ref}, i}$, ResidualKV encodes the difference between the current token and its reference in a low-dimensional latent space.
The nonlinear compressor is implemented as a SwiGLU projection:
\begin{displaymath}
  f_{\text{c}}(x) = (\text{Swish}(x\bm{W}_{c1}) \otimes x\bm{W}_{c2}) \bm{W}_{c3},
\end{displaymath}
where $\bm{W}_{c1}, \bm{W}_{c2}\in \mathbb{R}^{2d_{\text{k}}\times d_{\text{ff}}}$ and $\bm{W}_{c3}\in \mathbb{R}^{d_{\text{ff}}\times d_{\text{c}}}$; $\otimes$ denotes element-wise multiplication; $d_{\text{c}}$ is the compressed dimension; $d_{\text{ff}}$ is the SwiGLU intermediate width.
The residual code is:
\begin{displaymath}
  \bm{z}_{\Delta,i}
  = f_{\text{c}}(\bm{kv}_i) - f_{\text{c}}(\overline{\bm{KV}}_{\text{ref},i}),
\end{displaymath}
where $f_{\text{c}}: \mathbb{R}^{2d_{\text{k}}}\rightarrow \mathbb{R}^{d_{\text{c}}}$.
The resulting codes have shape $\bm{Z}_\Delta \in \mathbb{R}^{B\times L\times d_{\text{c}}}$. 
ResidualKV stores these compact codes together with the reference tokens required for reconstruction.

\paragraph{Residual Quantization}
Under tighter memory budgets, the residual codes are further quantized:
\begin{displaymath}
  \widetilde{\bm{z}}_{\Delta,i} = Q(\bm{z}_{\Delta,i}),
\end{displaymath}
where $Q(\cdot)$ denotes token-wise quantization and becomes the identity map when quantization is disabled. 
ResidualKV applies asymmetric min-max quantization independently to groups $\mathcal{G}$ of at most $g$ latent coordinates within each token:
\begin{displaymath}
  \bm{q}_{i,\mathcal{G}} = \operatorname{round}\!\Big(\frac{\bm{z}_{\Delta,i,\mathcal{G}} \!-\! z_{\min,\mathcal{G}}}{\Delta_{s,\mathcal{G}}}\Big), 
  \ \widetilde{\bm{z}}_{\Delta,i,\mathcal{G}} = \bm{q}_{i,\mathcal{G}}\Delta_{s,\mathcal{G}} \!+\! z_{\min,\mathcal{G}},
\end{displaymath}
where $\Delta_{s,\mathcal{G}} = \frac{z_{\max,\mathcal{G}}-z_{\min,\mathcal{G}}}{2^b-1}$; $b$ is the target bit-width; $z_{\min,\mathcal{G}}$ and $z_{\max,\mathcal{G}}$ are the group extrema. 
For a constant group, we set $Q(\bm{z}_{\Delta,i,\mathcal{G}})=\bm{z}_{\Delta,i,\mathcal{G}}$, and we use $g=32$ in all quantized configurations.
Because quantization follows reference subtraction, it operates on residuals rather than raw KV states.
Residualization suppresses outliers and concentrates values near zero (Figure~\ref{fig:combined_analysis_results}(f)), while the compressed codes remain low-magnitude and well distributed (Figure~\ref{fig:comp_kv_combined}), enabling aggressive low-bit quantization with limited accuracy degradation.

\paragraph{Flexible Reference Scheduling}
Because reference tokens are stored directly rather than residual-coded, their density directly determines storage overhead.
Thus, ResidualKV supports fixed-stride scheduling as a near-lossless default and dynamic-stride scheduling for ultra-long contexts.
Under fixed-stride scheduling, every $s$-th historical token is retained:
\begin{displaymath}
  \mathcal{T}_{\text{ref}, i} =
  \{ \bm{kv}_t \mid t<i,\ (t-1)\bmod s=0 \},
\end{displaymath}
This schedule distributes references uniformly and predictably, but their number grows linearly with context length.

\begin{figure}[t]
  \centering
  \includegraphics[width=0.99\linewidth]{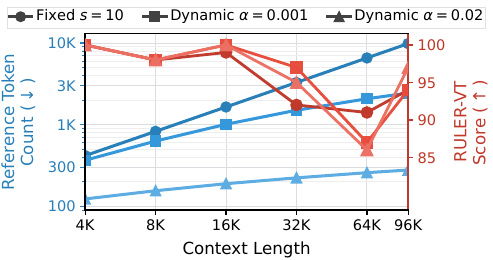}
  \caption{\textbf{Reference count and RULER-Variable Tracking (VT) performance under fixed- and dynamic-stride scheduling.}
  The left and right axes report the number of reference tokens and variable-tracking scores, respectively, under the same compressor and sparse-attention configuration.}
  \label{fig:dynamic_stride_reference_count_similarity}
  \vspace{-1.0em}
\end{figure}

To reduce this growth, dynamic-stride scheduling progressively increases the spacing between references.
Given an initial stride $s_0$, $n_{\mathrm{sink}}$ sink tokens, and a growth coefficient $\alpha$, reference positions follow:
\begin{displaymath}
  p_{m+1} = p_m + (s_0 + \lfloor \alpha(p_m-n_{\mathrm{sink}}) \rfloor).
\end{displaymath}
For $\alpha>0$ and sequence length $L$, the number of references is approximately:
\begin{displaymath}
  N_{\text{ref}}(L) \approx \log_{1+\alpha}\big(1 + \tfrac{\alpha(L-n_{\mathrm{sink}})}{s_0}\big),
\end{displaymath}
where $\alpha=0$ recovers fixed-stride scheduling. 
Dynamic stride reduces reference growth from $\mathcal{O}(L)$ to approximately $\mathcal{O}(\log L)$ without changing the residual compression, while preserving long-context performance (Figure~\ref{fig:dynamic_stride_reference_count_similarity}).
At 96K tokens, dynamic stride with $\alpha=0.02$ achieves a RULER-Variable-Tracking score~\cite{hsieh2024ruler} of 97.0 using only 278 references, versus 94.0 with 9,838 references for fixed stride.

\subsection{Residual Reconstruction}
\label{sec:method:reconstruction}

Once the residual codes and their references are stored, ResidualKV reconstructs only the attention-required KV states.

\paragraph{Lightweight Residual Reconstruction}
ResidualKV decodes a quantized residual code with a lightweight decompressor $f_{\text{d}}$, which is designed as a bias-free linear map:
\begin{displaymath}
  f_{\text{d}}(\bm{z}) = \bm{z}\bm{W}_{d},
\end{displaymath}
where $\bm{W}_{d} \in \mathbb{R}^{d_{\text{c}}\times 2d_{\text{k}}}$. For token $i$, the reconstructed residual and KV state are:
\begin{displaymath}
  \widehat{\bm{KV}}_{\Delta,i}=f_{\text{d}}(\widetilde{\bm{z}}_{\Delta,i}),
  \ \widehat{\bm{KV}}_i = \widehat{\bm{KV}}_{\Delta,i} + \overline{\bm{KV}}_{\text{ref},i}.
\end{displaymath}
The resulting vector is then separated and reshaped into key and value states for attention.

\begin{figure}[ht]
  \centering
  \includegraphics[width=0.96\linewidth]{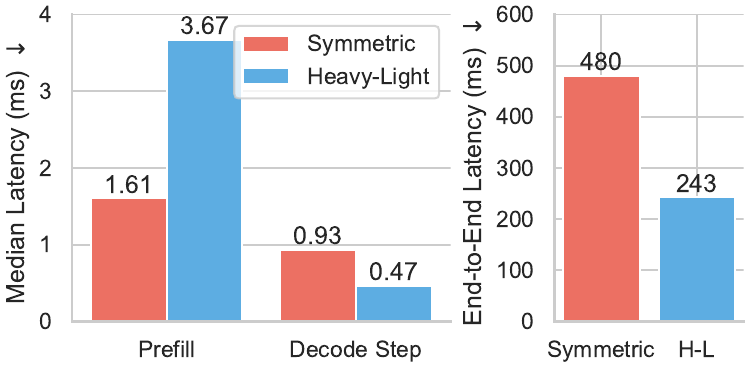}
  \caption{\textbf{Latency benefit of the heavy-compressor, lightweight-decompressor (heavy-light) design.}
  With comparable parameter counts (7.37M for the symmetric baseline and 7.34M for ours), the heavier one-time compressor increases prefill cost but reduces per-step reconstruction latency from 0.93 to 0.47 ms and total latency from 480 to 243 ms over a 4,096-token prefill and 512 decode steps.}
  \label{fig:lightweight_decompressor_latency}
  \vspace{-1.0em}
\end{figure}

This heavy-compressor, lightweight-decompressor design reflects the asymmetry of autoregressive inference.
Compression is performed only once when a token enters the cache and can be largely parallelized during prefill, whereas decompression may recur across many decoding steps, which are typically memory-bandwidth bound.
A single linear projection thus limits repeated weight loading and reconstruction overhead.
The bias term is omitted because the retrieved reference already provides the shared baseline, allowing $f_{\text{d}}$ to reconstruct only the token-specific deviation.
As shown in Figure~\ref{fig:lightweight_decompressor_latency}, this design reduces per-step reconstruction latency from 0.93 to 0.47 ms and the total reconstruction-path latency by 49.4\% on an H100.

\paragraph{Theoretical Justification}
We now connect the geometry of the actual $L_2$ retrieval rule to residual magnitude, group-wise quantization, and final KV reconstruction. All references below denote the effective numerical states used by both the residual encoder and decoder, including any storage round-trip. Let $\bm{\Delta}_i=\bm{kv}_i-\overline{\bm{KV}}_{\mathrm{ref},i}$. For the retrieval-induced pairs indexed by the data support $\mathcal{S}$, we define:
\begin{displaymath}
\begin{aligned}
  \eta_{\mathcal{S}} &:=  \sup_{\substack{i\in\mathcal{S}, \|\bm{\Delta}_i\|_2>0}}
  \frac{\|\bm{\Delta}_i-f_{\text{d}}(\bm{z}_{\Delta,i})\|_2}
  {\|\bm{\Delta}_i\|_2}, \\
  \gamma_{\mathcal{S}} &:= \sup_{\substack{i\in\mathcal{S}, \|\bm{\Delta}_i\|_2>0}}
  \frac{\|\bm{z}_{\Delta,i}\|_2}{\|\bm{\Delta}_i\|_2}.
\end{aligned}
\end{displaymath}
These quantities characterize the trained codec's relative reconstruction error and code amplification on retrieval-induced residuals, rather than on the full ambient KV space. If $\bm{\Delta}_i=0$, then $\bm{z}_{\Delta,i}=0$ and the bias-free decoder also returns zero, so the excluded case is reconstructed exactly.

\begin{theorem}
\label{thm:retrieval_distortion}
For a residual-coded token $i$, define the Root Mean Square (RMS) retrieval radius and within-reference dispersion as:
\begin{displaymath}
  \rho_i^2 = \frac{1}{k_i}\!\sum_{j\in\mathcal{R}_i}
  \|\bm{kv}_i \!-\! \bm{kv}_j\|_2^2, \ 
  V_i = \frac{1}{k_i}\!\sum_{j\in\mathcal{R}_i}
  \|\bm{kv}_j \!-\! \overline{\bm{KV}}_{\mathrm{ref},i}\|_2^2.
\end{displaymath}
Then the following exact decomposition holds:
\begin{displaymath}
  \rho_i^2=\|\bm{\Delta}_i\|_2^2+V_i.
\end{displaymath}
Let $M=2^b-1$ and $\bm{e}_{q,i}=Q(\bm{z}_{\Delta,i})-\bm{z}_{\Delta,i}$. For group-wise min-max quantization,
\begin{displaymath}
  \|\bm{e}_{q,i}\|_2 \le \frac{1}{2M}\!\sqrt{\sum_{\mathcal{G}}|\mathcal{G}|\, \operatorname{range}(\bm{z}_{\Delta,i,\mathcal{G}})^2}
  \le \frac{\sqrt{g/2}}{M} \|\bm{z}_{\Delta,i}\|_2.
\end{displaymath}
With $C_b = \frac{\|\bm{W}_d\|_{\mathrm{op}}\gamma_{\mathcal{S}}\sqrt{g/2}}{2^b-1}$,
the reconstructed KV state satisfies:
\begin{displaymath}
\begin{aligned}
  \textstyle \|\bm{kv}_i-\widehat{\bm{KV}}_i\|_2
  &\le \eta_{\mathcal{S}}\sqrt{\rho_i^2-V_i}
  +\|\bm{W}_d\|_{\mathrm{op}}
  \|\bm{e}_{q,i}\|_2 \\
  &\le (\eta_{\mathcal{S}}+C_b)\sqrt{\rho_i^2-V_i} \\
  &\le (\eta_{\mathcal{S}}+C_b)\rho_i.
\end{aligned}
\end{displaymath}
\end{theorem}

\begin{proof}
Writing $\overline{\bm{KV}}_{\mathrm{ref},i} = \tfrac{1}{k_i} \sum_{j \in \mathcal{R}_i} \bm{kv}_j$ and expanding each squared distance gives:
\begin{displaymath}
  \rho_i^2 
  = \frac{1}{k_i}\sum_{j\in\mathcal{R}_i} \|\bm{\Delta}_i+(\overline{\bm{KV}}_{\mathrm{ref},i}\!-\!\bm{kv}_j)\|_2^2 
  =\|\bm{\Delta}_i\|_2^2 \!+\! V_i,
\end{displaymath}
because the cross term vanishes after summing over the centered references. For any group $\mathcal{G}$, rounding changes each coordinate by at most $\operatorname{range}(\bm{z}_{\Delta,i,\mathcal{G}})/(2M)$, which gives the first quantization bound after summing squared errors. 

If $a$ and $b$ are the maximum and minimum coordinates in the group, then we have:
\begin{displaymath}
  \operatorname{range}(\bm{z}_{\Delta,i,\mathcal{G}})^2
  =(a-b)^2\le 2(a^2+b^2)
  \le 2\|\bm{z}_{\Delta,i,\mathcal{G}}\|_2^2.
\end{displaymath}
Using $|\mathcal{G}| \le g$ and summing over groups yields the second quantization bound. The definition of $\gamma_{\mathcal{S}}$ and the exact decomposition further give:
\begin{displaymath}
  \|\bm{z}_{\Delta,i}\|_2
  \le \gamma_{\mathcal{S}}\|\bm{\Delta}_i\|_2
  =\gamma_{\mathcal{S}}\sqrt{\rho_i^2-V_i}.
\end{displaymath}
Finally, linearity of $f_{\text{d}}(\bm{z})=\bm{z}\bm{W}_d$ gives:
\begin{displaymath}
\begin{aligned}
\bm{kv}_i-\widehat{\bm{KV}}_i
&=\bm{\Delta}_i-f_{\text{d}}(Q(\bm{z}_{\Delta,i})) \\
&=\bm{\Delta}_i-f_{\text{d}}(\bm{z}_{\Delta,i})
+f_{\text{d}}(\bm{z}_{\Delta,i}-Q(\bm{z}_{\Delta,i})).
\end{aligned}
\end{displaymath}
Applying the definition of $\eta_{\mathcal{S}}$, the decoder operator norm, and the two quantization bounds proves the result.
\end{proof}

If $f_{\text{c}}$ is $L_{\text{c}}$-Lipschitz on the corresponding KV--reference line segments, then $\gamma_{\mathcal{S}}\le L_{\text{c}}$, recovering a conventional Lipschitz-based sufficient bound. The data-dependent $\gamma_{\mathcal{S}}$ more directly characterizes the code amplification on the residual support.

For any size-$k_i$ subset $\mathcal{R}$ of the available reference pool, let
\begin{displaymath}
  \rho_i^2(\mathcal{R}) = \frac{1}{k_i} \sum_{\bm{kv}_j \in \mathcal{R}}
  \|\bm{kv}_i-\bm{kv}_j\|_2^2.
\end{displaymath}
Since ResidualKV selects the $k_i$ smallest squared $L_2$ distances,
\begin{displaymath}
  \mathcal{R}_i\in
  \operatorname*{arg\,min}_{\substack{\mathcal{R}\subseteq\mathcal{T}_{\mathrm{ref}}, |\mathcal{R}|=k_i}}
  \rho_i^2(\mathcal{R}).
\end{displaymath}
Thus, for a fixed support-wise coefficient $\eta_{\mathcal{S}}+C_b$, the actual retrieval rule minimizes the retrieval-radius term in the coarse distortion bound among all size-$k_i$ candidate subsets. Theorem~\ref{thm:retrieval_distortion} therefore shows that ResidualKV turns geometric proximity to globally retrieved references into an end-to-end reconstruction guarantee: both residual magnitude and quantization-induced distortion scale with the RMS retrieval radius. The identity separates the centroid residual from within-reference dispersion; it does not imply that increasing dispersion improves compression across different reference sets.

\begin{corollary}
\label{cor:shared_component}
Suppose $\bm{kv}_i=\bm{s}_{c_i}+\bm{\epsilon}_i$, where $\mathbb{E}[\bm{\epsilon}_i\mid c_i]=0$ and $\operatorname{Cov}(\bm{\epsilon}_i\mid c_i)=\bm{\Sigma}_{c_i}$. If the $k_i$ compatible references share component $c_i$ and their perturbations are mutually independent and independent of $\bm{\epsilon}_i$, then
\begin{displaymath}
\begin{aligned}
  \mathbb{E}[\bm{\Delta}_i\mid c_i]&=0, \\
  \operatorname{Cov}(\bm{\Delta}_i\mid c_i)
  &=\big(1+\tfrac{1}{k_i}\big)\bm{\Sigma}_{c_i}, \\
  \mathbb{E}[\|\bm{\Delta}_i\|_2^2\mid c_i]
  &=\big(1+\tfrac{1}{k_i}\big)\operatorname{Tr}(\bm{\Sigma}_{c_i}).
\end{aligned}
\end{displaymath}
\end{corollary}

\begin{proof}
Compatible reference subtraction gives $\bm{\Delta}_i=\bm{\epsilon}_i-k_i^{-1}\sum_{j=1}^{k_i}\bm{\epsilon}_{r_j}$, exactly canceling $\bm{s}_{c_i}$. Conditional zero mean and independence then give the stated moments.
\end{proof}

Under this idealized compatible-reference model, the stochastic interpretation is motivated by the shared geometric structure of contextual representations~\cite{ethayarajh2019contextual}. It also explains the benefit and saturation of top-$k_i$ averaging: the reference-noise contribution decreases from $\bm{\Sigma}_{c_i}$ to $\bm{\Sigma}_{c_i}/k_i$, so the total residual covariance moves from $2\bm{\Sigma}_{c_i}$ toward $\bm{\Sigma}_{c_i}$. Because the reducible term decays only as $1/k_i$, additional references provide diminishing variance reduction and may introduce weaker matches, consistent with Figure~\ref{fig:all_ablations}.

\subsection{Joint Sparse Inference}
\label{sec:method:sparse_inference}

\begin{figure}[t]
  \centering
  \includegraphics[width=0.99\linewidth]{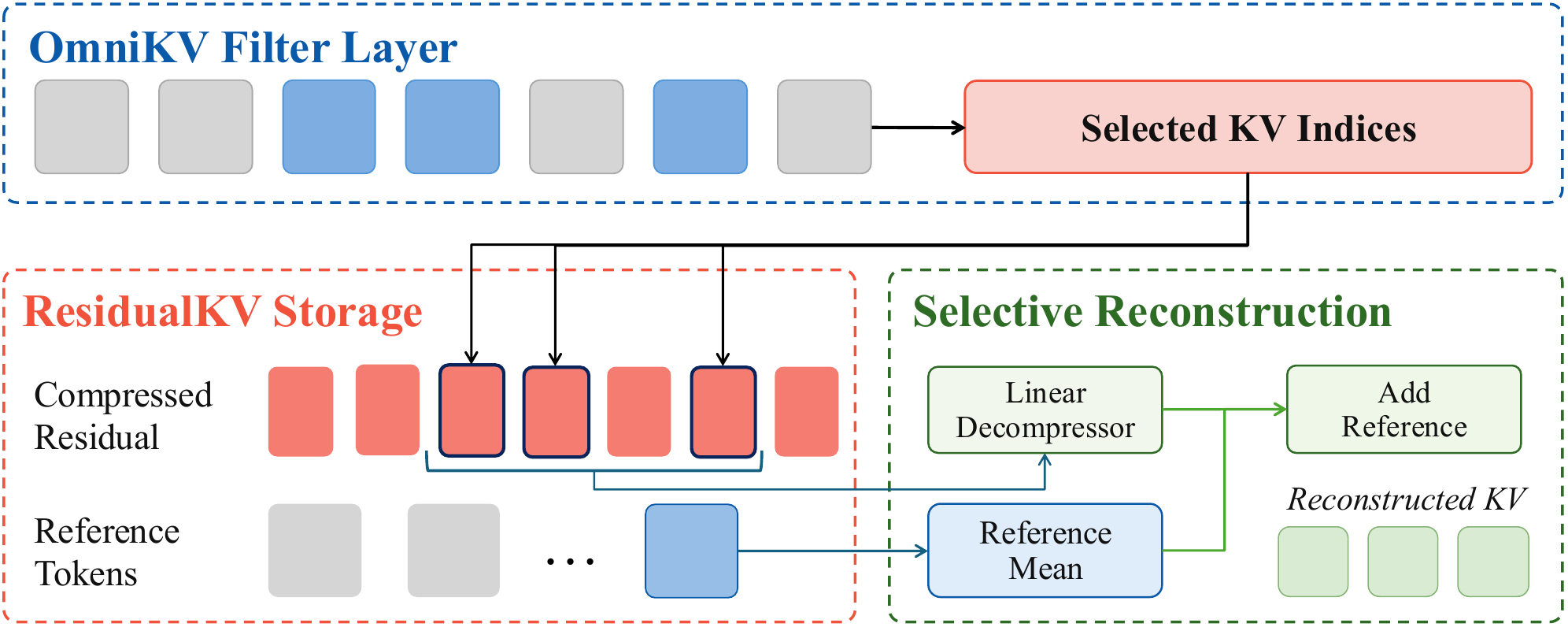}
  \caption{\textbf{Joint sparse inference with OmniKV and ResidualKV.}
  An OmniKV filter layer selects historical token indices using full attention. ResidualKV then fetches and reconstructs only the corresponding residual codes through lightweight linear decompression and reference addition.}
  \label{fig:omnikv_with_reskv}
  \vspace{-1.0em}
\end{figure}

ResidualKV can independently reconstruct the complete compressed cache. Its main inference benefit, however, arises with sparse selection, which allows only active KV states to be fetched and reconstructed, as illustrated in Figure \ref{fig:omnikv_with_reskv}.

\paragraph{Selected KV Reconstruction}
We primarily integrate ResidualKV with OmniKV~\cite{hao2025omnikv}. 
OmniKV designates a small set of \emph{filter layers} $\mathbb{L}_{\text{filter}}$, whose full-attention scores select important historical tokens for the remaining sparse layers $\mathbb{L}_{\text{sparse}}$. 
For a decoding step $t$ and a filter layer $l\in\mathbb{L}_{\text{filter}}$, OmniKV computes attention scores over the full historical cache:
\begin{displaymath}
  \bm{a}_{l,t} = \operatorname{softmax}\Big(\frac{\bm{q}_{l,t}\bm{K}_{l,<t}^{\top}}{\sqrt{D}}\Big), 
  \ \mathcal{I}_{t} = \underset{j<t}{\operatorname{arg\,top\,-}k_{\text{sparse}}}\; \bm{a}_{l,t,j},
\end{displaymath}
where $\mathcal{I}_{t}$ contains the $k_{\text{sparse}}$ selected indices.
ResidualKV replaces the raw sparse-layer history with reference tokens and residual codes, reconstructing only states in $\mathcal{I}_{t}$:
\begin{displaymath}
  \widehat{\bm{kv}}_{l,j}
  =
  f_{\text{d}}(\widetilde{\bm{z}}_{\Delta,l,j})
  + \overline{\bm{KV}}_{R,l,j},
\end{displaymath}
where $j \in \mathcal{I}_{t}$. After separating the reconstructed vectors into keys and values, sparse attention is computed over this subset:
\begin{displaymath}
  \bm{o}_{l,t} = \operatorname{softmax}\Big(
  \frac{\bm{q}_{l,t}\widehat{\bm{K}}_{l,\mathcal{I}_{t}}^{\top}}{\sqrt{D}}
  \Big)
  \widehat{\bm{V}}_{l,\mathcal{I}_{t}},
\end{displaymath}
where $l \in \mathbb{L}_{\text{sparse}}$.
OmniKV thus determines which tokens are active, while ResidualKV stores and reconstructs their states. 
Together, they restrict each sparse layer to $|\mathcal{I}_{t}|=k_{\text{sparse}}$ tokens and decode only the corresponding residual codes, reducing both attention computation and memory traffic.

\paragraph{Remarks}
ResidualKV also supports methods that directly retain per-layer token subsets.
Let $\mathcal{S}^{(m)}_{l,t}$ denote the tokens retained by selection or eviction method $m$ at layer $l$ and decoding step $t$. 
ResidualKV augments this subset with the references required for reconstruction:
\begin{displaymath}
  \textstyle \mathcal{U}^{(m)}_{l,t} =
  \mathcal{S}^{(m)}_{l,t} \cup \mathcal{T}^{\text{ref}}_{l,t}.
\end{displaymath}
Tokens in $\mathcal{T}^{\text{ref}}_{l,t}$ remain uncompressed as reconstruction bases, while those in $\mathcal{S}^{(m)}_{l,t} \setminus \mathcal{T}^{\text{ref}}_{l,t}$ are stored as residual codes and reconstructed on demand.
This directly supports the layer-wise subsets retained by SnapKV~\cite{li2024snapkv} and KVzip~\cite{kim2025kvzip}.

ResidualKV is thus orthogonal to the token-selection policy. 
OmniKV derives dynamic masks from full filter-layer states, while ResidualKV compresses sparse-layer states and reconstructs only selected tokens; with SnapKV or KVzip, it further compresses retained subsets. 
These combinations jointly reduce attention FLOPs, KV-cache storage, and memory traffic.

\subsection{Hybrid Objective Training}
\label{sec:method:training}

Efficient selective reconstruction requires both numerical fidelity and preservation of prediction-relevant information.
Because reconstruction error alone does not account for the unequal downstream importance of different latent directions, ResidualKV jointly optimizes Mean Squared Error (MSE) reconstruction and Next-Token Prediction (NTP) losses.

\paragraph{MSE Reconstruction Loss}
The reconstruction loss encourages recovered KV states to match their uncompressed targets:
\begin{displaymath}
  \mathcal{L}_\text{rec} = \sum_{l,t} \|\bm{KV}_{l,t} - \widehat{\bm{KV}}_{l,t}\|_2^2.
  \vspace{-0.5em}
\end{displaymath}
This objective preserves the geometry and token-level variations of the original KV representations.

\paragraph{NTP Loss}
MSE treats all deviations equally, despite their unequal effects on attention and predictions; errors in sensitive directions can alter attention logits, be amplified by softmax, and propagate across layers. 
ResidualKV therefore adds task-aware NTP supervision, computed for $\mathcal{T}=(x_1,\cdots,x_T)$ from logits using reconstructed KV states:
\begin{displaymath}
  \mathcal{L}_\text{ntp}(\phi;\theta)
  = -\frac{1}{T-1}\sum_{t=1}^{T-1}
  \log p_{\theta,\phi}(x_{t+1}\mid x_{\leq t}; \widehat{\bm{KV}}_\phi).
\end{displaymath}
where $\theta$ denotes frozen LLM parameters and $\phi$ denotes the learnable ResidualKV modules. 

\paragraph{Overall Objective}
The overall objective is:
\begin{displaymath}
  \mathcal{L} = \mathcal{L}_\text{rec} + \mathcal{L}_\text{ntp}(\phi;\theta).
\end{displaymath}
Both losses update only $\phi$ through the frozen backbone $\theta$: $\mathcal{L}_\text{rec}$ preserves numerical fidelity, while $\mathcal{L}_\text{ntp}$ retains latent directions most consequential to the model's predictions.

\section{Experiments}
\label{sec:expt}

\subsection{Experimental Setup}
\label{sec:expt:setup}

\paragraph{Models and Training Configuration}
We evaluate ResidualKV on multimodal and text-only benchmarks.
For the former, we cover LLaVA-OneVision-7B (\textbf{LLaVA-OV-7B}) \cite{li2024llava} and Qwen3-VL-8B-Instruct (\textbf{Qwen3-VL-8B-Ins}) \cite{bai2025qwen3}, covering LLaVA- and Qwen-style backbones;
for the latter, we use Llama-3.1-8B-Instruct (\textbf{Llama3.1-8B-Ins}) \cite{grattafiori2024llama3}, Qwen2.5-7B-Instruct-1M (\textbf{Qwen2.5-7B-Ins}) \cite{yang2025qwen2_1m}, Qwen2.5-32B-Instruct (\textbf{Qwen2.5-32B-Ins}) \cite{team2024qwen2}, Qwen3-4B-Instruct-2507 (\textbf{Qwen3-4B-Ins}), and Qwen3-4B-Thinking-2507 (\textbf{Qwen3-4B-Think}) \cite{yang2025qwen3}.

For all models, we freeze the pretrained backbone and optimize only the ResidualKV modules.
Unless otherwise specified, we use AdamW with a learning rate of $2 \!\times\! 10^{-4}$, batch size 1, one gradient-accumulation step, 2\% linear warmup, and linear decay to zero.
For multimodal models, ResidualKV compresses the language-decoder KV states without pruning visual tokens. 
For Qwen3-VL-8B-Ins, we freeze the vision encoder, multimodal projector, and language decoder and train only the compressor and decompressor on streamed LLaVA-OV data for 30k steps with a learning rate of $5\times10^{-5}$.

Text models with 7B/8B parameters are trained on 160M FineWeb-Edu tokens~\cite{penedo2024fineweb}, packed into 8,192-token sequences, requiring approximately \textbf{8 GPU hours} and \textbf{\$8}; Qwen2.5-32B-Ins requires 14 GPU hours. Training uses \texttt{PyTorch} and \texttt{transformers} \cite{wolf2020transformers} on one NVIDIA RTX PRO 6000 (Blackwell), with \texttt{bitsandbytes} and gradient checkpointing~\cite{gradient_ckpt} additionally enabled for Qwen2.5-32B-Ins.

\begin{table}[t]
\centering
\small
\renewcommand{\arraystretch}{1.2}
\caption{\textbf{ResidualKV configuration by model.} $2d_{\text{k}}$ is the original per-token KV dimension, $d_{\text{c}}$ is the compressed residual dimension, $d_{\text{ff}}$ is the SwiGLU intermediate width, and $L_{\text{full}}$ lists the full-attention layers used.}
\label{tab:full_attn_layers_config}
\resizebox{\linewidth}{!}{%
\begin{tabular}{lrrrr}
\toprule
\textbf{Model} & $\bm{2d_{\text{k}}}$ & $\bm{d_{\text{c}}}$ & $\bm{d_{\text{ff}}}$ & $\bm{L_{\text{full}}}$ \\
\midrule
\textbf{LLaVA-OV-7B}      & 1,024 & 384 & 3,072 & 0,1,4,7,14\\
\textbf{Qwen3-VL-8B-Ins}  & 2,048 & 384 & 3,072 & 0,1,2,3,10,21,29 \\
\midrule
\textbf{Llama-3.1-8B-Ins} & 2,048 & 512 & 3,072 & 0,1,2,8,18 \\
\textbf{Qwen2.5-7B-Inst}  & 1,024 & 384 & 3,072 & 0,1,2,4,7,14 \\
\textbf{Qwen2.5-32B-Ins}  & 2,048 & 512 & 4,096 & 0,1,2,3,4,5,17,29,40 \\
\textbf{Qwen3-4B-Ins}     & 2,048 & 512 & 3,072 & 0,3,9,13,16,21,28 \\
\textbf{Qwen3-4B-Think}   & 2,048 & 512 & 3,072 & 0,3,9,13,16,21,28 \\
\bottomrule
\end{tabular}}
\end{table}
\setlength{\textfloatsep}{1.0em}

\paragraph{Compression and Layer Configuration}
ResidualKV uses the asymmetric SwiGLU compressor and bias-free linear decompressor described in Section~\ref{sec:method:compression}.
The compressed residual dimension $d_{\mathrm{c}}$ and intermediate width $d_{\mathrm{ff}}$ are configured per model (Table~\ref{tab:full_attn_layers_config}). 
Unless otherwise stated, retrieval uses the top-$k\!=\!4$ nearest references from a fixed-stride reference set with $s\!=\!10$, i.e., around a $10\%$ reference keep ratio.

We use a hybrid configuration in which a small subset of layers retains the complete history for full attention, while the remaining layers apply ResidualKV with sparse attention.
Sparse-layer residual codes are quantized to 4 bits by default and to \texttt{INT2} for matched-budget multimodal experiments, aligning their compute ratio with the 10\% budget of visual-token-reduction baselines.

\begin{table*}[t]
\centering
\small
\setkrcrwidths{10}{100}
\renewcommand{\arraystretch}{1.3}
\caption{\textbf{Main results on multimodal benchmarks.}
We evaluate ResidualKV on LLaVA-OV-7B and Qwen3-VL-8B-Ins backbones across seven representative multimodal benchmarks.}
\label{tab:multimodal_perf}
\resizebox{1.0\textwidth}{!}{
\begin{tabular}{c l>{\columncolor[HTML]{EBF5FF}}c ccccccc >{\columncolor[HTML]{FBE3E1}}c}
\toprule
\textbf{Model} & \textbf{Method} & (\textbf{KR}, \textbf{CR}) $\downarrow$ & \textbf{MMB $\uparrow$} & \textbf{MME $\uparrow$} & \textbf{MMMU $\uparrow$} & \textbf{POPE $\uparrow$} & \textbf{SQA $\uparrow$} & \textbf{V-MME $\uparrow$} & \textbf{MVB $\uparrow$} & \textbf{Average $\uparrow$} \\
\midrule
\multirow{5}{*}{\rotatebox[origin=c]{90}{\shortstack{\textit{LLaVA-OV}\\\textit{-7B}}}} & \textbf{FastV}~\cite{chen2024image_fastv}       & \krcr{10}{10} & 80.3 & 81.7/1816.0 & 44.9 & 76.8 & 83.3 & 53.6 & 51.0 & 67.4 \\
& \textbf{PACT}~\cite{dhouib2025pact}             & \krcr{10}{10} & 84.9 & 84.7/2013.9 & 47.8 & 88.7 & 92.3 & 58.7 & 53.1 & 72.9 \\
& \textbf{DivPrune}~\cite{alvar2025divprune}      & \krcr{10}{10} & 80.4 & 79.6/1745.2 & 46.4 &85.4 & 83.5 & 54.7 & 52.1 & 68.9 \\
& \textbf{FastVID}~\cite{shen2025fastvid}         & \krcr{10}{10} & -- & -- & -- & -- & -- & 56.3 & 53.5 & -- \\
& \textbf{ResidualKV (Ours)}                         & \textbf{\krcr{8}{10}} & 85.1 & 84.9/2032.3 & 48.6 & 89.2 & 95.2 & 58.6 & 53.7 & \textbf{73.6} \\
\midrule
\multirow{3}{*}{\rotatebox[origin=c]{90}{\scriptsize\shortstack{\textit{Qwen3-VL}\\\textit{-8B-Ins}}}} & \textbf{FastV}~\cite{chen2024image_fastv}       & \krcr{10}{10} & 72.4 & 80.0/1641.6 & 47.7 & 74.2 & 78.7 & 47.4 & 48.4 & 64.1 \\
& \textbf{DivPrune}~\cite{alvar2025divprune}      & \krcr{10}{10} & 79.4 & 84.8/1877.0 & 49.1 & 83.6 & 79.8 & 47.4 & 46.7 & 67.3 \\
& \textbf{ResidualKV (Ours)}                         & \textbf{\krcr{10}{10}} & 88.9 & 89.3/2399.7 & 52.9 & 88.8 & 94.5 & 51.4 & 55.2 & \textbf{74.4} \\
\bottomrule
\end{tabular}}
\end{table*}
\setlength{\textfloatsep}{1.0em}

\paragraph{Baselines}
For multimodal evaluation, we compare \textbf{ResidualKV} with \textbf{FastV}~\cite{chen2024image_fastv}, \textbf{PACT}~\cite{dhouib2025pact}, \textbf{DivPrune}~\cite{alvar2025divprune}, and \textbf{FastVID}~\cite{shen2025fastvid}.
These methods reduce visual tokens through pruning, clustering, diversity preservation, or adaptive token-density control, whereas ResidualKV compresses the language-decoder KV cache throughout decoding.

For text-only evaluation, we compare three families:
(i) static eviction methods that permanently discard selected tokens, including \textbf{SnapKV}~\cite{li2024snapkv}, \textbf{PyramidKV}~\cite{cai2025pyramidkv}, \textbf{AdaKV}~\cite{feng2025adakv}, and \textbf{KVzip}~\cite{kim2025kvzip}; 
(ii) dynamic sparsity methods that select tokens per query while retaining the full cache, including \textbf{Quest}~\cite{tang2024quest} and \textbf{OmniKV}~\cite{hao2025omnikv}; and 
(iii) eviction-free compression methods, namely \textbf{KIVI}~\cite{liu2024kivi} for low-bit quantization and \textbf{Palu}~\cite{chang2025palu} for low-rank projection without exploiting cross-token similarity.

\paragraph{Benchmarks and Evaluation Protocols}
Multimodal evaluation spans seven benchmarks: 
MMBench (\textbf{MMB}) \cite{liu2024mmbench} and \textbf{MME} \cite{fu2026mme} for general perception and cognition; 
\textbf{MMMU}~\cite{yue2024mmmu} and ScienceQA (\textbf{SQA})~\cite{lu2022learn} for knowledge-intensive visual reasoning;
\textbf{POPE}~\cite{li2023evaluating} for object hallucination; and 
Video-MME (\textbf{V-MME})~\cite{fu2025videomme} and MVBench (\textbf{MVB})~\cite{li2024mvbench} for video and temporal understanding.

Text-only evaluation uses \textbf{LongBench}~\cite{bai2024longbench}, \textbf{SCBench}~\cite{li2025scbench}, and \textbf{AIME}~\cite{maa_aime}, covering single-turn long-context understanding, same-context multi-query interaction, and long-form mathematical reasoning, respectively.
Following \textbf{AdaKV}~\cite{feng2025adakv}, we assess all 16 LongBench datasets, spanning single- and multi-document QA, summarization, few-shot learning, synthetic tasks, and code generation.
ResidualKV uses dynamic stride with $\alpha\!=\!0.001$ on LongBench and fixed stride with $\alpha\!=\!0$ on SCBench.
Due to computational constraints, SCBench evaluation uses representative same-context, different-query (\textbf{SCDQ}) tasks covering string and semantic retrieval, global in-context information, and multi-task summarization with needle-in-a-haystack evaluation. 
AIME permits up to 32,768 output tokens for long reasoning traces.

\paragraph{Memory and Compute Metrics}
We report the KV cache Keep Ratio (\textbf{KR}), defined as GPU-resident KV memory relative to a full FP16 cache, and the Compute Ratio (\textbf{CR}), defined as the fraction of historical tokens used in attention.
Methods are matched by a target ratio $r \in [0,1]$, allowing token budgets to scale with context length; the tables report these ratios as percentages.
For group-wise $b$-bit quantization with group size 32, the relative storage cost is
$q_b=b/16+2/32$, accounting for the quantized payload and two FP16 extrema per group.
Hence, $(\text{KR}, \text{CR})$ is $(r, r)$ for static eviction (SnapKV, PyramidKV), $(100\%, r)$ for dynamic selection (OmniKV, Quest), $(q_b, 100\%)$ for KIVI, and $(d_{\text{low}}/d_{\text{orig}}, 100\%)$ for Palu.

For ResidualKV, let $L=L_{\text{full}}+L_{\text{sparse}}$, where $L_{\text{full}}$ layers use full attention and $L_{\text{sparse}}$ use compressed sparse attention. 
Full-attention layers use 4-bit KV quantization, while sparse layers store FP8 references and $b$-bit residual codes, with $b\!=\!2$ for matched-budget multimodal evaluation and $b\!=\!4$ otherwise. 
The resulting ratios are:
\begin{displaymath}
  \text{KR}=\frac{L_{\text{full}}}{L} q_4 \!+\! \frac{L_{\text{sparse}}}{L} \big(\frac{1}{2} \rho_{\text{ref}} \!+\! q_b\frac{d_{\mathrm{c}}}{2d_{\mathrm{k}}}\big),
  \ \text{CR}=\frac{L_{\text{full}}}{L} \!+\! \frac{L_{\text{sparse}}}{L} r.
\end{displaymath}
where $\rho_{\text{ref}}$ is the reference-token keep ratio, $d_{\mathrm{c}}$ and $2d_{\mathrm{k}}$ are the compressed and original per-token KV dimensions, and $1/2$ reflects FP8 storage relative to FP16.
For the default 4-bit fixed-stride Llama-3.1-8B setting ($L=32$, $\rho_{\text{ref}}=0.10$, and $d_{\mathrm{c}}/(2d_{\mathrm{k}})=0.25$), KR is 15.7\%/15.1\% for 5/4 full-attention and 27/28 sparse layers, decreasing to 13.4\%/13.0\% under dynamic stride ($\alpha=0.001$) on LongBench.

\subsection{Main Performance}
\label{sec:expt:performance}

\begin{table*}[t]
\centering
\small
\setkrcrwidths{100}{100}
\renewcommand{\arraystretch}{1.2}
\caption{\textbf{Main results on LongBench.} We compare ResidualKV against state-of-the-art baselines across varying model scales. Variants marked $^\dagger$ omit the neural
compressor; all methods are matched by target ratio $r$ and averaged over 16 datasets.}
\label{tab:longbench_perf}
\resizebox{1.0\linewidth}{!}{
\begin{tabular}{c l >{\columncolor[HTML]{EBF5FF}}c cccccc >{\columncolor[HTML]{FBE3E1}}c}
\toprule
\textbf{Model} & \textbf{Method}  & (\textbf{KR}, \textbf{CR}) $\downarrow$ &
\textbf{Single-Doc $\uparrow$} &
\textbf{Multi-Doc $\uparrow$} &
\textbf{Summ. $\uparrow$} &
\textbf{Few-Shot $\uparrow$} &
\textbf{Synthetic $\uparrow$} &
\textbf{Code $\uparrow$} &
\textbf{Average $\uparrow$} \\
\midrule
\multirow{9}{*}{\rotatebox[origin=c]{90}{\shortstack{\textit{Llama3.1}\\\textit{-8B-Ins}}}} & \textbf{SnapKV}~\cite{li2024snapkv}                 & \krcr{30}{30} & 43.6 & 46.3 & 26.4 & 68.4 & 55.4 & 60.2 & 50.1 \\
& \textbf{PyramidKV}~\cite{cai2025pyramidkv}          & \krcr{30}{30} & 43.7 & 46.6 & 26.2 & 68.8 & 55.6 & 59.7 & 50.1 \\
& \textbf{Quest}~\cite{tang2024quest}                 & \krcr{100}{30} & 42.6 & 46.0 & 29.1 & 69.1 & 54.8 & 57.8 & 49.9 \\
& \textbf{KIVI-2bit}~\cite{liu2024kivi}               & \krcr{19}{100} & 43.2 & 46.3 & 28.9 & 69.1 & 54.8 & 59.9 & 50.3 \\
& \textbf{AdaKV}~\cite{feng2025adakv}                 & \krcr{30}{30} & 43.1 & 46.0 & 26.4 & 69.5 & 55.0 & 58.2 & 49.7 \\
& \textbf{OmniKV}~\cite{hao2025omnikv}                & \krcr{100}{30} & 42.9 & 46.0 & 28.5 & 68.7 & 54.9 & 60.0 & 50.2 \\
& \textbf{KVzip}~\cite{kim2025kvzip}                  & \krcr{30}{30} & 43.2 & 42.7 & 28.9 & 68.2 & 53.0 & 61.8 & 49.6 \\
& \textbf{ResidualKV$^\dagger$ (Ours)}                   & \krcr{27}{30} & 43.7 & 46.7 & 28.7 & 69.1 & 54.6 & 59.6 & \textbf{50.4} \\
& \textbf{ResidualKV (Ours)}                             & \textbf{\krcr{13}{30}} & 43.2 & 47.0 & 27.9 & 69.3 & 54.4 & 59.8 & 50.3\\

\midrule
\multirow{8}{*}{\rotatebox[origin=c]{90}{\shortstack{\textit{Qwen2.5}\\\textit{-7B-Ins}}}} & \textbf{SnapKV}~\cite{li2024snapkv}                 & \krcr{30}{30} & 42.0 & 48.9 & 26.7 & 68.9 & 54.3 & 42.5 & 47.2 \\
& \textbf{PyramidKV}~\cite{cai2025pyramidkv}          & \krcr{30}{30} & 41.8 & 49.1 & 25.7 & 69.2 & 54.3 & 41.7 & 47.0 \\
& \textbf{KIVI-2bit}~\cite{liu2024kivi}               & \krcr{19}{100} & 41.2 & 48.7 & 28.1 & 69.3 & 52.5 & 42.0 & 47.0 \\
& \textbf{OmniKV}~\cite{hao2025omnikv}                & \krcr{100}{30} & 41.5 & 49.0 & 28.2 & 69.1 & 53.8 & 41.6 & 47.2 \\
& \textbf{Palu}~\cite{chang2025palu}                  & \krcr{50}{100} & 34.4 & 35.6 & 27.5 & 68.7 & 45.0 & 21.3 & 38.8 \\
& \textbf{KVzip}~\cite{kim2025kvzip}                  & \krcr{30}{30} & 42.0 & 49.3 & 26.9 & 67.2 & 53.3 & 45.7 & 47.4 \\
& \textbf{ResidualKV$^\dagger$ (Ours)}                   & \krcr{26}{30} & 41.2 & 49.6 & 28.3 & 68.4 & 53.8 & 42.9 & \textbf{47.4} \\
& \textbf{ResidualKV (Ours)}                             & \textbf{\krcr{16}{30}} & 41.9 & 49.4 & 28.2 & 68.6 & 53.5 & 42.2 & 47.3\\

\midrule
\multirow{5}{*}{\rotatebox[origin=c]{90}{\shortstack{\textit{Qwen2.5}\\\textit{-32B-Ins}}}} & \textbf{SnapKV}~\cite{li2024snapkv}                 & \krcr{30}{30} & 41.9& 54.4 & 25.5 & 67.4 & 55.8 & 42.1 & 47.8 \\
& \textbf{PyramidKV}~\cite{cai2025pyramidkv}          & \krcr{30}{30} & 41.2 & 54.0 & 24.5 & 67.3 & 56.0 & 40.6 & 47.3 \\
& \textbf{KIVI-2bit}~\cite{liu2024kivi}               & \krcr{19}{100} & 42.3 & 53.7 & 26.6 & 68.5 & 54.9 & 44.4 & 48.4 \\
& \textbf{OmniKV}~\cite{hao2025omnikv}                & \krcr{100}{30} & 42.8 & 54.9 & 27.4 & 67.2 &55.8 & 43.0 & \textbf{48.5} \\
& \textbf{ResidualKV (Ours)}                             & \textbf{\krcr{13}{30}} & 41.3 & 51.6 & 26.7 & 69.4 & 58.5 & 42.7 & 48.4\\

\midrule
\multirow{4}{*}{\rotatebox[origin=c]{90}{\shortstack{\textit{Qwen3}\\\textit{-4B-Ins}}}} & \textbf{SnapKV}~\cite{li2024snapkv}                 & \krcr{30}{30} & 40.2 & 42.1 & 25.4 & 66.1 & 50.4 & 58.1 & 47.1 \\
& \textbf{PyramidKV}~\cite{cai2025pyramidkv}          & \krcr{30}{30} & 37.2 & 38.1 & 24.9 & 61.2 & 29.5 & 56.2 & 41.2 \\
& \textbf{OmniKV}~\cite{hao2025omnikv}                & \krcr{100}{30} & 40.4 & 42.8 & 25.8 & 68.9 & 50.4 & 59.4 & 47.9 \\
& \textbf{ResidualKV (Ours)}                             & \textbf{\krcr{13}{30}} & 40.5 & 43.0 & 25.7 & 68.3 & 50.5 & 59.5 & \textbf{47.9} \\
\bottomrule
\end{tabular}}
\end{table*}

\paragraph{Multimodal Benchmarks}
Table~\ref{tab:multimodal_perf} shows that ResidualKV preserves near-full performance across seven image and video benchmarks.
At a matched \textbf{10\% CR}, the average changes only from 73.8 to 73.6 on LLaVA-OV-7B and from 74.5 to 74.4 on Qwen3-VL-8B-Ins. 
ResidualKV also outperforms all visual-token-reduction baselines by up to \textbf{7.1 points} on Qwen3-VL-8B-Ins while using only \textbf{8\%--10\% KR}.
These results show that the Long-Range Inter-Token Similarity and Smooth Residual Distribution identified in Section~\ref{sec:observations} also characterize multimodal KV representations.

\paragraph{Single-Turn LongBench}
Table~\ref{tab:longbench_perf} demonstrates consistent performance across four backbones ranging from 4B to 32B parameters.
ResidualKV achieves the best or tied-best average among compressed methods in most settings and remains within 0.1 points elsewhere, using only \textbf{13\%--16\% KR} and \textbf{30\% CR}.
This corresponds to up to \textbf{7$\times$} less KV storage than dynamic-selection methods, \textbf{2$\times$} less than static eviction at 30\% KR, and over \textbf{3$\times$} less attention computation.

Static methods such as SnapKV and KVzip degrade more under tight budgets because discarded tokens cannot be recovered, whereas dynamic methods such as OmniKV preserve accuracy but retain the full cache. 
ResidualKV combines the advantages of both approaches by reducing the stored cache and the active attention set.

\begin{table}[t]
\centering
\small
\setlength{\tabcolsep}{2pt}
\setkrcrwidths{20}{100}
\renewcommand{\arraystretch}{1.3}
\caption{\textbf{Main results on SCBench.} R.KV, R.PS, En.QA, S+N, and MS denote key-value retrieval, prefix-suffix retrieval, English question answering, summarization plus needle-in-a-haystack retrieval, and many-shot in-context learning.}
\label{tab:scbench_perf}
\resizebox{\linewidth}{!}{%
\begin{tabular}{c l >{\columncolor[HTML]{EBF5FF}}c ccccc >{\columncolor[HTML]{FBE3E1}}c}
\toprule
\textbf{Model} & \textbf{Method} & (\textbf{KR}, \textbf{CR}) $\downarrow$ & \textbf{R.KV $\uparrow$} & \textbf{R.PS $\uparrow$} & \textbf{En.QA $\uparrow$} & \textbf{S+N $\uparrow$} & \textbf{MS $\uparrow$} & \textbf{Avg. $\uparrow$} \\
\midrule
\multirow{4}{*}{\rotatebox[origin=c]{90}{\shortstack{\textit{Qwen2.5}\\\textit{7B-Ins}}}} & \textbf{SnapKV}         & \krcr{20}{20} & 1.4  & 1.6 & 18.0 & 28.6 & 59.3 & 22.5 \\
& \textbf{KIVI-2bit}      & \krcr{19}{100} & 0.0 & 0.0 & 25.6 & 58.9 & 59.3 & 24.4 \\
& \textbf{KVzip}          & \krcr{20}{20} & 36.2  & 3.0 & 44.0 & 66.5 & 33.7 & 36.7 \\
& \textbf{ResidualKV}        & \textbf{\krcr{19}{20}} & 52.6 & 23.4 & 29.9 & 47.6 & 61.5 & \textbf{40.7} \\
\midrule
\multirow{3}{*}{\rotatebox[origin=c]{90}{\shortstack{\textit{Qwen3}\\\textit{4B-Ins}}}} & \textbf{SnapKV}         & \krcr{20}{20} & 2.6  & 0.0 & 16.8 & 50.1 & 61.1 & 22.8 \\
& \textbf{KIVI-2bit}      & \krcr{19}{100} & 0.2 & 0.2 & 16.5 & 55.4 & 53.0 & 20.7 \\
& \textbf{ResidualKV}        & \textbf{\krcr{16}{20}} & 34.6 & 0.0 & 26.8 & 51.1 & 67.0 & \textbf{32.9} \\
\bottomrule
\end{tabular}}
\end{table}
\setlength{\textfloatsep}{1.0em}

\paragraph{Multi-Turn SCBench}
Table~\ref{tab:scbench_perf} shows that ResidualKV achieves the highest compressed-method average on both backbones (\textbf{40.7} and \textbf{32.9}), surpassing the strongest baseline by \textbf{4.0} and \textbf{10.1 points}, respectively.
This advantage indicates that its residual representation preserves information that may become relevant in later turns. 
The margin is particularly large on key-value retrieval, where ResidualKV scores 52.6 versus 36.2 for KVzip and 1.4 for SnapKV, highlighting the vulnerability of irreversible eviction.


\paragraph{Long-Form Reasoning on AIME}
Table \ref{tab:aime_perf} shows that ResidualKV fully preserves the full-cache score of Qwen3-4B-Think (\textbf{80.0}), outperforming SnapKV and OmniKV by \textbf{20.0} and \textbf{3.3 points}, respectively.
Thus, compressing historical KV states preserves the long reasoning trajectories required for mathematical problem solving.

\begin{table}[t]
\centering
\small
\setlength{\tabcolsep}{5pt}
\renewcommand{\arraystretch}{1.0}
\caption{\textbf{Main results on the AIME reasoning benchmark.}}
\label{tab:aime_perf}
\begin{tabular}{lrrrr}
\toprule
\textbf{Model}& \textbf{Full}& \textbf{SnapKV}& \textbf{OmniKV}& \textbf{ResidualKV}\\
\midrule
\textit{Qwen3-4B-Think} & 80.0 & 60.0 & 76.7 & \textbf{80.0} \\
\bottomrule
\end{tabular}
\end{table}

\begin{figure}[t]
  \centering
  \includegraphics[width=0.95\linewidth]{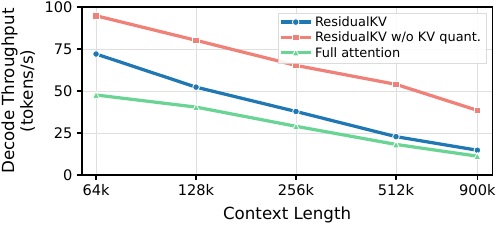}
  \caption{\textbf{Batch-size-1 decoding throughput across context lengths.} Comparison of full attention and ResidualKV with and without KV-cache quantization.}
  \label{fig:decode_tp}
  \vspace{-0.5em}
\end{figure}

\paragraph{Inference Throughput}
Figure~\ref{fig:decode_tp} reports batch-size-1 decoding throughput on one NVIDIA H100 GPU.
ResidualKV outperforms full attention at all evaluated context lengths by up to \textbf{1.5$\times$} higher throughput in its standard quantized configuration and \textbf{3.4$\times$} without KV-cache quantization.
The gap arises mainly because quantized full-attention layers must reconstruct the complete KV cache from KIVI-style low-bit representations before attention, introducing substantial overhead.

\subsection{Design Analysis}
\label{sec:expt:analysis}

\paragraph{Dynamic-Stride Scheduling}
Dynamic-stride scheduling limits reference-memory growth by progressively increasing the spacing between stored references.
At a context length of 262K, increasing $\alpha$ from 0 to 0.001 reduces the reference set by approximately \textbf{8$\times$}, from 26,189 to 3,354 tokens (9.99\% to 1.28\%), without changing the LongBench average of 47.4.
Performance remains within \textbf{0.4 points} even at $\alpha=0.1$.
Figure~\ref{fig:dynamic_stride_reference_count_similarity} and Table~\ref{tab:dynamic_stride_summary} therefore validate the near-logarithmic reference growth derived in Section~\ref{sec:method:compression}.

\begin{table}[!t]
\centering
\small
\setlength{\tabcolsep}{4pt}
\renewcommand{\arraystretch}{1.1}
\caption{\textbf{Dynamic stride trade-off on Qwen2.5-7B-Ins.} Increasing $\alpha$ sharply reduces the reference set at 262K context with a modest change in the LongBench average score.}
\label{tab:dynamic_stride_summary}
\begin{tabular}{lrrr}
\toprule
$\bm{\alpha}$ & \textbf{$\bm{N_{\mathrm{ref}}}$ @ 262K $\downarrow$} & \textbf{Ref. Fraction $\downarrow$} & \textbf{LongBench $\uparrow$} \\
\midrule
0.000 & 26,189 & 9.99\% & \textbf{47.4} \\
0.001 &  3,354 & 1.28\% & 47.4 \\
0.020 &    319 & 0.12\% & 47.3 \\
0.050 &    149 & 0.06\% & 47.0 \\
0.100 & \textbf{83} & \textbf{0.03\%} & 47.1 \\
\bottomrule
\end{tabular}%
\end{table}

\begin{figure}[ht]
  \centering
  \begin{minipage}[b]{0.51\columnwidth}
    \centering
    \includegraphics[width=\linewidth]{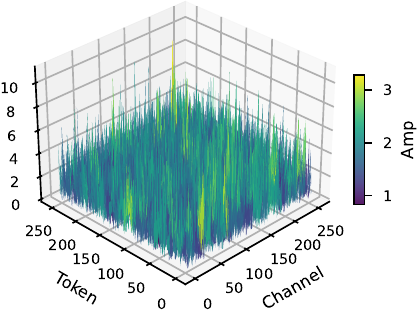}
    \par{\footnotesize (a)\par}
  \end{minipage}
  \hfill
  \begin{minipage}[b]{0.45\columnwidth}
    \centering
    \includegraphics[width=\linewidth]{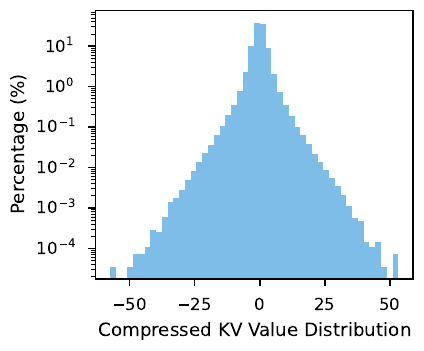}
    \par{\footnotesize (b)\par}
  \end{minipage}
  \caption{\textbf{Distribution of compressed KV representations.} (a) Magnitudes remain smooth across tokens and channels without localized spikes. (b) Values concentrate near zero, facilitating low-bit quantization.}
  \label{fig:comp_kv_combined}
  \vspace{-0.5em}
\end{figure}

\paragraph{Compatibility with Quantization}
The main experiments use ResidualKV's quantized deployment configuration: 4-bit KV states in full-attention layers and token-wise quantized residual codes $\bm{z}_{\Delta}$ in sparse layers.
Figure~\ref{fig:comp_kv_combined} explains the robustness of this configuration using compressed representations from Qwen2.5-7B-Ins.
Their magnitudes vary smoothly across tokens and channels without prominent spikes, while their values concentrate strongly near zero. 
This distribution is well suited to low-bit quantization, explaining the near-full accuracy in Tables~\ref{tab:longbench_perf} and~\ref{tab:scbench_perf} despite the substantial reduction in KR.

\paragraph{Standalone Use and Compositionality}
Table~\ref{tab:standalone_composition} verifies the operating modes described in Section~\ref{sec:method:sparse_inference}.
Standalone ResidualKV reconstructs the complete compressed cache and matches the full-cache average of \textbf{50.5} at \textbf{16\% KR}. When composed with OmniKV or SnapKV, it retains averages of \textbf{50.3} and \textbf{50.1} at \textbf{13\%} and \textbf{17.5\% KR}, respectively. 
ResidualKV can therefore operate independently or complement dynamic and static token-selection policies.

\begin{table}[t]
\centering
\small
\setkrcrwidths{17.5}{100}
\renewcommand{\arraystretch}{1.2}
\caption{\textbf{Standalone and compositional use of ResidualKV on LongBench with Llama-3.1-8B-Ins.} Standalone ResidualKV reconstructs all compressed KV pairs without an external sparsity mask, while ResidualKV+OmniKV and ResidualKV+SnapKV compose residual compression with dynamic and static token selection, respectively.}
\label{tab:standalone_composition}
\begin{tabular}{l >{\columncolor[HTML]{EBF5FF}}c >{\columncolor[HTML]{FBE3E1}}r}
\toprule
\textbf{Method} & (\textbf{KR}, \textbf{CR}) $\downarrow$ &
\textbf{Average $\uparrow$} \\
\midrule
\textbf{ResidualKV Standalone} & \krcr{16}{100} & \textbf{50.5} \\
\textbf{ResidualKV + OmniKV}   & \textbf{\krcr{13}{30}} & 50.3 \\
\textbf{ResidualKV + SnapKV}   & \krcr{17.5}{40} & 50.1 \\
\bottomrule
\end{tabular}%
\end{table}

\paragraph{Train Short, Test Long}
Although trained on only 8,192-token sequences, ResidualKV generalizes to contexts beyond 100K tokens (Table~\ref{tab:scbench_perf}). 
We attribute this length generalization to performing reference retrieval and compression on pre-RoPE KV representations, making the learned transformation largely independent of absolute token position.

\begin{figure}[ht]
  \centering
  \vspace{-0.5em}
  \includegraphics[width=0.99\linewidth]{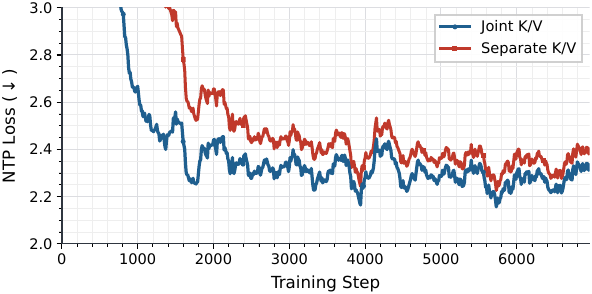}
  \caption{\textbf{Joint vs. separate K/V compression.} Step-aligned NTP loss under matched trainable parameters, compression ratios, and training steps, smoothed over 200 steps.}
  \label{fig:kv_concat_ntp_loss}
  \vspace{-0.5em}
\end{figure}

\paragraph{Joint K/V Compression}
ResidualKV concatenates keys and values before reference retrieval and residual compression.
Figure~\ref{fig:kv_concat_ntp_loss} compares this joint design with separate K/V compressors under matched parameter counts, compression ratios, and training steps.
Joint compression converges faster and maintains lower NTP loss over most of training, suggesting that keys and values share underlying hidden-state structure that is captured more effectively by a unified compressor.

\subsection{Ablation Study}
\label{sec:expt:ablation}

\paragraph{Component Ablation}
We ablate ResidualKV's two core components on Llama3.1-8B-Ins ($2d_{\text{k}}\!=\!2{,}048$): reference-based residual construction and learned residual compression.
To isolate their effects, we disable additional KV quantization and increase $d_{\mathrm{c}}$ from 512 to 768, yielding a comparable keep ratio of approximately 35\%.
As shown in Table~\ref{tab:ablation_module}, removing the compressor-decompressor pair $(f_c,f_d)$ lowers the average by \textbf{3.6 points}, while removing the reference tokens $\mathcal{T}_{\mathrm{ref}}$ lowers it by \textbf{4.4 points}. 
Both residualization and learned compression are therefore essential.

\begin{table}[t]
\centering
\small
\setlength{\tabcolsep}{1.2pt}
\setkrcrwidths{47}{300}
\renewcommand{\arraystretch}{1.4}
\caption{\textbf{Component-wise ablation study on Llama-3.1-8B-Ins.} SD, MD, Sum, FS, Syn, and
Code denote single-document QA, multi-document QA, summarization, few-shot learning, synthetic tasks, and code generation; KV quantization is disabled and $d_{\mathrm{c}}\!=\!768$ so that all KRs are comparable.}
\label{tab:ablation_module}
\resizebox{0.99\linewidth}{!}{
\begin{tabular}{l >{\columncolor[HTML]{EBF5FF}}c cccccc >{\columncolor[HTML]{FBE3E1}}c}
\toprule
\textbf{Method}  & (\textbf{KR}, \textbf{CR}) $\downarrow$ & \textbf{SD $\uparrow$} & \textbf{MD $\uparrow$} & \textbf{Sum $\uparrow$} & \textbf{FS $\uparrow$} & \textbf{Syn $\uparrow$} & \textbf{Code $\uparrow$} & \textbf{Avg. $\uparrow$} \\
\midrule
\textbf{ResidualKV}                                & \krcr{45}{30} & 43.2 & 47.1 & 27.9 & 69.2 & 54.4 & 59.8 & \textbf{50.3} \\
\quad \textbf{w/o $f_{\mathrm{c}}$ and $f_{\mathrm{d}}$}              & \krcr{45}{30} & 34.4 & 43.2 & 24.5 & 65.9 & 55.1 & 57.2 & 46.7 \\
\quad \textbf{w/o Ref. Tokens $\mathcal{T}$}    & \krcr{47}{30} & 39.3 & 45.3 & 22.1 & 62.3 & 50.8 & 55.4 & 45.9 \\
\bottomrule
\end{tabular}}
\end{table}

\begin{figure*}[t]
  \centering
  \includegraphics[width=0.99\textwidth]{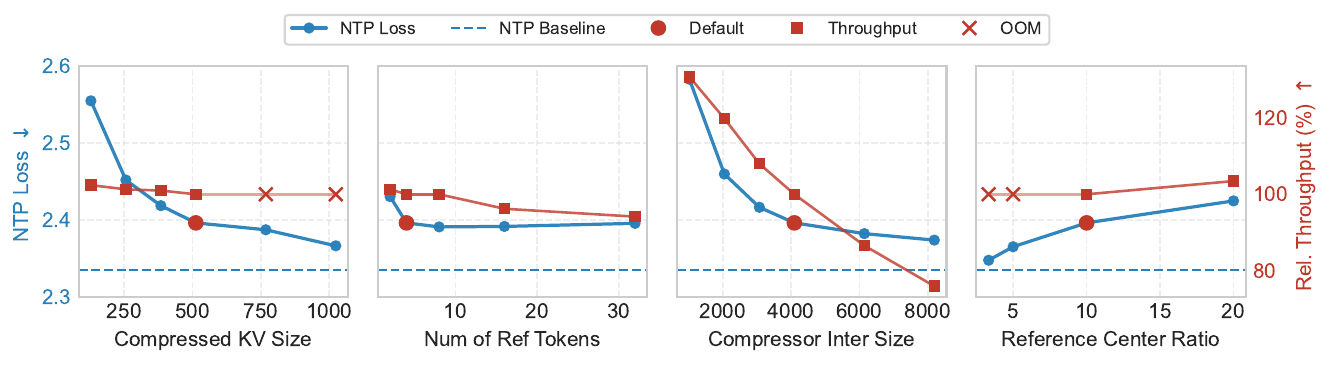}
  \vspace{-1.0em}
  \caption{\textbf{Hyperparameter sensitivity.} We vary the compressed dimension, number of references, compressor intermediate width, and reference stride while holding other settings fixed. Blue/red curves show NTP loss/relative decoding throughput, red circles denote the defaults, and red crosses indicate OOM.}
  \label{fig:all_ablations}
  \vspace{-0.5em}
\end{figure*}

\begin{figure*}[t]
  \centering
  \includegraphics[width=0.99\textwidth]{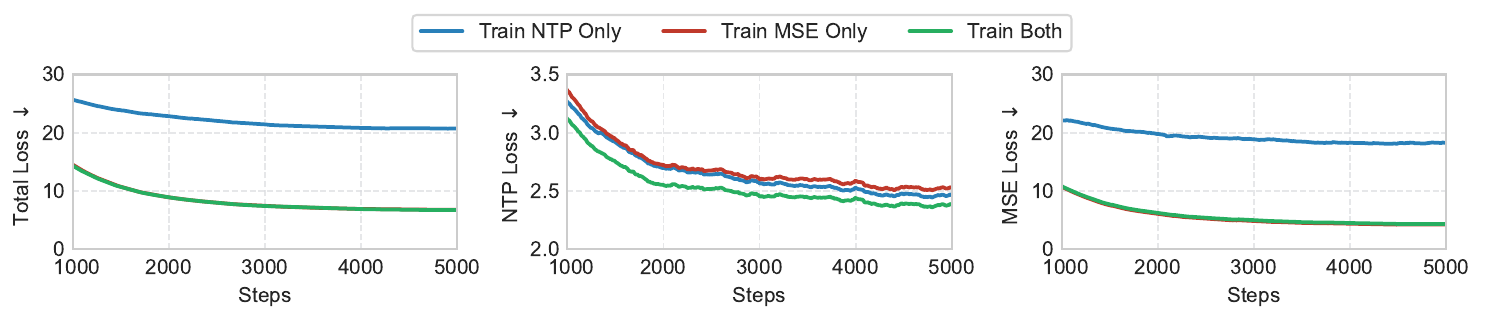}
  \vspace{-1.0em}
  \caption{\textbf{Ablation of the hybrid training objective.} Comparison of NTP-only, MSE-only, and joint training using total, NTP, and MSE losses. Joint training preserves prediction quality while minimizing reconstruction error.}
  \label{fig:ablation_on_ntp_mse}
  \vspace{-0.5em}
\end{figure*}

\paragraph{Hyperparameter Sensitivity}
Figure~\ref{fig:all_ablations} examines the quality-efficiency trade-offs of four hyperparameters by varying each while holding the others fixed. Throughput is measured at a context length of 128K and batch size 16.
\begin{itemize}[nolistsep, left=0pt]
  \item \textbf{Compressed Dimension ($d_{\mathrm{c}}$):} Increasing $d_{\mathrm{c}}$ reduces NTP loss but yields diminishing returns, consumes more memory, and eventually causes out-of-memory (OOM).

  \item \textbf{Number of References ($k$):} Additional references initially reduce reference noise, but the benefit quickly saturates as less compatible references are included.
  Values of $k=4$ or $8$ therefore offer a favorable quality-efficiency balance, consistent with Section~\ref{sec:method:compression}.

  \item \textbf{Compressor Intermediate Width ($d_{\mathrm{ff}}$):} Increasing $d_{\mathrm{ff}}$ improves prediction quality but adds parameter and computational overhead, reducing decoding throughput.

  \item \textbf{Reference Stride ($s$):} Decreasing $s$ retains more reference tokens and improves prediction quality but increases storage, retrieval, and reconstruction costs, reducing throughput and potentially causing OOM.
\end{itemize}
Overall, the default settings lie near the knee of each trade-off curve, providing strong prediction quality without excessive memory or computation.

\begin{figure*}[t]
  \centering
  \includegraphics[width=0.95\textwidth]{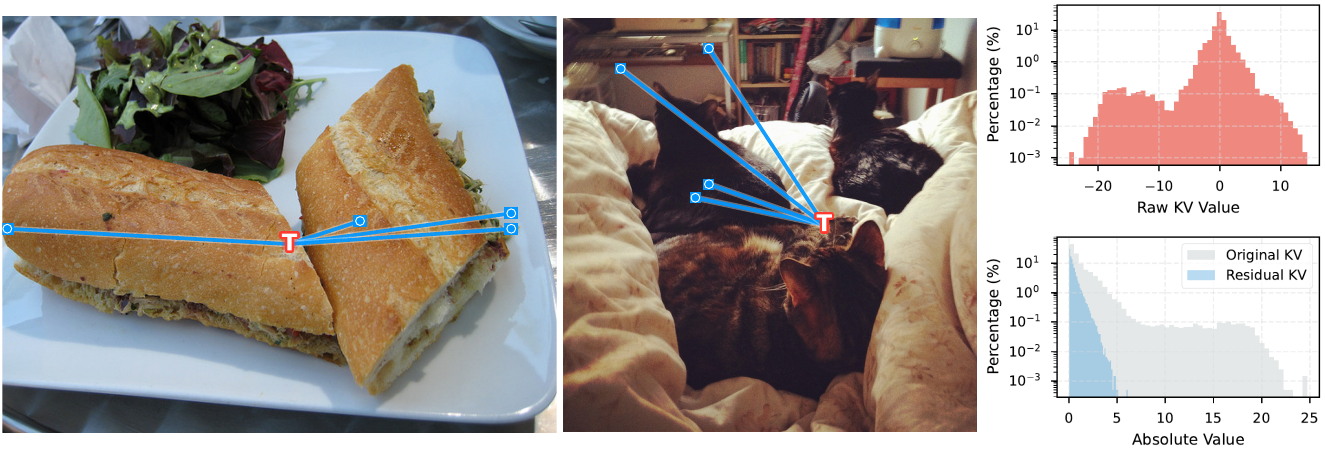}
  \vspace{-0.5em}
  \caption{\textbf{Multimodal reference retrieval and residual distribution.} 
  Left and middle: target tokens $\bm{T}$ on object boundaries retrieve references from semantically and structurally similar boundaries elsewhere in the image. Right: reference subtraction transforms the long-tailed raw KV distribution into a compact, near-zero residual distribution.}
  \label{fig:multimodal_case_study}
  \vspace{-0.75em}
\end{figure*}

\paragraph{Hybrid-Objective Ablation}
Figure~\ref{fig:ablation_on_ntp_mse} compares NTP-only, MSE-only, and joint training.
MSE-only training also lowers NTP loss, indicating that numerical fidelity benefits language modeling. Conversely, NTP-only training does not reduce MSE, showing that prediction-oriented supervision alone does not enforce uniform numerical fidelity across KV dimensions. 
Joint optimization achieves both low reconstruction error and strong NTP performance, validating the complementary objectives introduced in Section~\ref{sec:method:training}.

\subsection{Case Study}
\label{sec:expt:case_study}

Figure~\ref{fig:multimodal_case_study} illustrates why reference-based residualization remains effective for multimodal inputs.
In the left and middle examples, target tokens $\bm{T}$ on structural boundaries, such as a sandwich edge or cat-ear contour, retrieve references from similar boundaries elsewhere in the image, including other sandwich edges, another cat's ear, and a box edge. These long-range correspondences show that semantically and structurally related visual tokens share substantial information in their KV representations.
The right panel further shows that subtracting this shared component transforms the broad, long-tailed raw KV distribution into a compact, near-zero residual distribution.
The case study therefore directly visualizes the Long-Range Inter-Token Similarity and Smooth Residual Distribution identified in Section~\ref{sec:observations}, explaining why multimodal residual states are amenable to aggressive compression and low-bit quantization.

\section{Conclusions}
\label{sec:conclusions}

We introduce \textbf{ResidualKV}, which reframes KV-cache compression as a reference--residual factorization rather than independent compression or eviction of token states. 
Our analysis reveals that distant tokens often share substantial KV structure and that subtracting globally similar references produces compact, smooth residuals. 
ResidualKV exploits these properties by retaining reusable references and encoding only token-specific residuals, thereby preserving historical information while supporting quantization and selective reconstruction. 
It can operate independently or complement static and dynamic token-selection methods to jointly reduce cache storage, attention computation, and memory traffic.
Across text-only and multimodal backbones, ResidualKV maintains near-full performance with \textbf{8\%--16\%} of the full KV storage, including on multi-turn interaction and long-form reasoning, and improves batch-size-1 decoding throughput by up to \textbf{1.5$\times$}. 
Dynamic-stride scheduling further limits reference growth at ultra-long contexts without degrading accuracy. 
These findings establish global cross-token redundancy as a useful design principle for KV-cache optimization and suggest that preserving shared structure with compact token-specific deviations offers a promising path toward scalable long-context inference.



\bibliography{reference}

@inproceedings{ethayarajh2019contextual,
  title={How contextual are contextualized word representations? Comparing the geometry of BERT, ELMo, and GPT-2 embeddings},
  author={Ethayarajh, Kawin},
  booktitle={EMNLP},
  pages={55--65},
  year={2019},
  url={https://aclanthology.org/D19-1006.pdf}
}

@inproceedings{tian2024drivevlm,
  title={{DriveVLM: The Convergence of Autonomous Driving and Large Vision-Language Models}},
  author={Xiaoyu Tian and Junru Gu and Bailin Li and Yicheng Liu and Yang Wang and Zhiyong Zhao and Kun Zhan and Peng Jia and XianPeng Lang and Hang Zhao},
  booktitle={CoRL},
  pages={4698--4726},
  year={2024},
  url={https://openreview.net/forum?id=928V4Umlys}
}

@inproceedings{fu2025videomme,
  title={{Video-MME: The First-Ever Comprehensive Evaluation Benchmark of Multi-modal LLMs in Video Analysis}},
  author={Fu, Chaoyou and Dai, Yuhan and Luo, Yongdong and Li, Lei and Ren, Shuhuai and Zhang, Renrui and Wang, Zihan and Zhou, Chenyu and Shen, Yunhang and Zhang, Mengdan and others},
  booktitle={CVPR},
  pages={24108--24118},
  year={2025},
  url={https://ieeexplore.ieee.org/abstract/document/11093290}
}

@inproceedings{liu2024mmbench,
  title={{MMBench: Is Your Multi-modal Model an All-Around Player?}},
  author={Liu, Yuan and Duan, Haodong and Zhang, Yuanhan and Li, Bo and Zhang, Songyang and Zhao, Wangbo and Yuan, Yike and Wang, Jiaqi and He, Conghui and Liu, Ziwei and others},
  booktitle={ECCV},
  pages={216--233},
  year={2024},
  url={https://link.springer.com/chapter/10.1007/978-3-031-72658-3_13}
}

@inproceedings{fu2026mme,
  title={{MME: A Comprehensive Evaluation Benchmark for Multimodal Large Language Models}},
  author={Fu, Chaoyou and Chen, Peixian and Shen, Yunhang and Qin, Yulei and Zhang, Mengdan and Lin, Xu and Yang, Jinrui and Zheng, Xiawu and Li, Ke and Sun, Xing and others},
  booktitle={NeurIPS},
  year={2026},
  url={https://openreview.net/forum?id=DgH9YCsqWm},
}

@inproceedings{yue2024mmmu,
  title={{MMMU: A Massive Multi-discipline Multimodal Understanding and Reasoning Benchmark for Expert AGI}},
  author={Yue, Xiang and Ni, Yuansheng and Zhang, Kai and Zheng, Tianyu and Liu, Ruoqi and Zhang, Ge and Stevens, Samuel and Jiang, Dongfu and Ren, Weiming and Sun, Yuxuan and others},
  booktitle={CVPR},
  pages={9556--9567},
  year={2024},
  url={https://ieeexplore.ieee.org/abstract/document/10656299}
}

@inproceedings{li2023evaluating,
  title={Evaluating Object Hallucination in Large Vision-Language Models},
  author={Li, Yifan and Du, Yifan and Zhou, Kun and Wang, Jinpeng and Zhao, Xin and Wen, Ji-Rong},
  booktitle={EMNLP},
  pages={292--305},
  year={2023},
  url={https://aclanthology.org/2023.emnlp-main.20/}
}

@article{lu2022learn,
  title={Learn to Explain: Multimodal Reasoning via Thought Chains for Science Question Answering},
  author={Lu, Pan and Mishra, Swaroop and Xia, Tanglin and Qiu, Liang and Chang, Kai-Wei and Zhu, Song-Chun and Tafjord, Oyvind and Clark, Peter and Kalyan, Ashwin},
  journal={NeurIPS},
  volume={35},
  pages={2507--2521},
  year={2022},
  url={https://openreview.net/forum?id=HjwK-Tc_Bc}
}

@inproceedings{li2024mvbench,
  title={{MVBench: A Comprehensive Multi-modal Video Understanding Benchmark}},
  author={Li, Kunchang and Wang, Yali and He, Yinan and Li, Yizhuo and Wang, Yi and Liu, Yi and Wang, Zun and Xu, Jilan and Chen, Guo and Luo, Ping and Wang, Limin and Qiao, Yu},
  booktitle={CVPR},
  pages={22195--22206},
  year={2024},
  url={https://ieeexplore.ieee.org/abstract/document/10658165}
}

@article{wu2024longvideobench,
  title={{LongVideoBench: A Benchmark for Long-context Interleaved Video-Language Understanding}},
  author={Wu, Haoning and Li, Dongxu and Chen, Bei and Li, Junnan},
  journal={NeurIPS},
  volume={37},
  pages={28828--28857},
  year={2024},
  url={https://openreview.net/forum?id=3G1ZDXOI4f}
}

@inproceedings{li2025scbench,
  title={{SCBench: A KV Cache-Centric Analysis of Long-Context Methods}},
  author={YUCHENG LI and Huiqiang Jiang and Qianhui Wu and Xufang Luo and Surin Ahn and Chengruidong Zhang and Amir H. Abdi and Dongsheng Li and Jianfeng Gao and Yuqing Yang and Lili Qiu},
  booktitle={ICLR},
  year={2025},
  volume={2025},
  pages={66063--66093},
  url={https://openreview.net/forum?id=gkUyYcY1W9}
}

@article{zhang2025pqcache,
  title={{PQCache: Product Quantization-based KVCache for Long Context LLM Inference}},
  author={Zhang, Hailin and Ji, Xiaodong and Chen, Yilin and Fu, Fangcheng and Miao, Xupeng and Nie, Xiaonan and Chen, Weipeng and Cui, Bin},
  journal={Proceedings of the ACM on Management of Data (SIGMOD)},
  volume={3},
  number={3},
  pages={1--30},
  year={2025},
  url={https://dl.acm.org/doi/abs/10.1145/3725338}
}

@inproceedings{li2024snapkv,
  title={{SnapKV: LLM Knows What You are Looking for Before Generation}},
  author={Yuhong Li and Yingbing Huang and Bowen Yang and Bharat Venkitesh and Acyr Locatelli and Hanchen Ye and Tianle Cai and Patrick Lewis and Deming Chen},
  booktitle={NeurIPS},
  year={2024},
  volume={37},
  pages={22947--22970},
  url={https://openreview.net/forum?id=poE54GOq2l}
}

@inproceedings{hao2025omnikv,
  title={{OmniKV: Dynamic Context Selection for Efficient Long-Context LLMs}},
  author={Jitai Hao and Yuke Zhu and Tian Wang and Jun Yu and Xin Xin and Bo Zheng and Zhaochun Ren and Sheng Guo},
  booktitle={ICLR},
  year={2025},
  volume={2025},
  pages={87443--87464},
  url={https://openreview.net/forum?id=ulCAPXYXfa}
}

@inproceedings{zhang2023h2o,
  title={{H$_2$O: Heavy-Hitter Oracle for Efficient Generative Inference of Large Language Models}},
  author={Zhenyu Zhang and Ying Sheng and Tianyi Zhou and Tianlong Chen and Lianmin Zheng and Ruisi Cai and Zhao Song and Yuandong Tian and Christopher Re and Clark Barrett and Zhangyang Wang and Beidi Chen},
  booktitle={NeurIPS},
  volume={36},
  pages={34661--34710},
  year={2023},
  url={https://openreview.net/forum?id=RkRrPp7GKO}
}

@article{grattafiori2024llama3,
  title={{The Llama 3 Herd of Models}},
  author={Grattafiori, Aaron and Dubey, Abhimanyu and Jauhri, Abhinav and Pandey, Abhinav and Kadian, Abhishek and Al-Dahle, Ahmad and Letman, Aiesha and Mathur, Akhil and Schelten, Alan and Vaughan, Alex and others},
  journal={arXiv preprint arXiv:2407.21783},
  year={2024},
  url={https://arxiv.org/abs/2407.21783}
}

@inproceedings{tang2024quest,
  title={{QUEST: Query-Aware Sparsity for Efficient Long-Context {LLM} Inference}},
  author={Jiaming Tang and Yilong Zhao and Kan Zhu and Guangxuan Xiao and Baris Kasikci and Song Han},
  booktitle={ICML},
  pages={47901--47911},
  year={2024},
  url={https://openreview.net/forum?id=KzACYw0MTV}
}

@inproceedings{oren2024transformers_tova,
  title={{Transformers are Multi-State RNNs}},
  author={Oren, Matanel and Hassid, Michael and Yarden, Nir and Adi, Yossi and Schwartz, Roy},
  booktitle={EMNLP},
  pages={18724--18741},
  year={2024},
  url={https://aclanthology.org/2024.emnlp-main.1043/}
}

@article{liu2023scissorhands,
  title={{Scissorhands: Exploiting the Persistence of Importance Hypothesis for LLM KV Cache Compression at Test Time}},
  author={Liu, Zichang and Desai, Aditya and Liao, Fangshuo and Wang, Weitao and Xie, Victor and Xu, Zhaozhuo and Kyrillidis, Anastasios and Shrivastava, Anshumali},
  journal={NeurIPS},
  volume={36},
  pages={52342--52364},
  year={2023},
  url={https://openreview.net/forum?id=JZfg6wGi6g}
}

@inproceedings{liu2024intactkv,
  title={{IntactKV: Improving Large Language Model Quantization by Keeping Pivot Tokens Intact}},
  author={Liu, Ruikang and Bai, Haoli and Lin, Haokun and Li, Yuening and Gao, Han and Xu, Zhengzhuo and Hou, Lu and Yao, Jun and Yuan, Chun},
  booktitle={Findings of ACL},
  pages={7716--7741},
  year={2024},
  url={https://aclanthology.org/2024.findings-acl.460/}
}

@article{lin2025quantization,
  title={{Quantization Meets dLLMs: A Systematic Study of Post-training Quantization for Diffusion LLMs}},
  author={Lin, Haokun and Xu, Haobo and Wu, Yichen and Guo, Ziyu and Zhang, Renrui and Lu, Zhichao and Wei, Ying and Zhang, Qingfu and Sun, Zhenan},
  journal={arXiv preprint arXiv:2508.14896},
  year={2025},
  url={https://arxiv.org/abs/2508.14896}
}

@inproceedings{guo2025enhancing,
  title={{Enhancing RAG Efficiency with Adaptive Context Compression}},
  author={Guo, Shuyu and Zhang, Shuo and Ren, Zhaochun},
  booktitle={Findings of EMNLP},
  pages={24061--24076},
  year={2025},
  url={https://aclanthology.org/2025.findings-emnlp.1307/},
}

@inproceedings{xiao2024efficient,
  title={{Efficient Streaming Language Models with Attention Sinks}},
  author={Xiao, Guangxuan and Tian, Yuandong and Chen, Beidi and Han, Song and Lewis, Mike},
  booktitle={ICLR},
  volume={2024},
  pages={21875--21895},
  year={2024},
  url={https://openreview.net/forum?id=NG7sS51zVF}
}

@article{liu2024deepseekv2,
  title={{DeepSeek-V2: A Strong, Economical, and Efficient Mixture-of-Experts Language Model}},
  author={Liu, Aixin and Feng, Bei and Wang, Bin and Wang, Bingxuan and Liu, Bo and Zhao, Chenggang and Dengr, Chengqi and Ruan, Chong and Dai, Damai and Guo, Daya and others},
  journal={arXiv preprint arXiv:2405.04434},
  year={2024},
  url={https://arxiv.org/abs/2405.04434}
}

@inproceedings{hao2025tokenLRC,
  title={{A Token is Worth over 1,000 Tokens: Efficient Knowledge Distillation through Low-Rank Clone}},
  author={Jitai Hao and Qiang Huang and Hao Liu and Xinyan Xiao and Zhaochun Ren and Jun Yu},
  booktitle={NeurIPS},
  volume={38},
  pages={53678--53710},
  year={2025},
  url={https://openreview.net/forum?id=LVDRJE4xQ2}
}

@article{zhang2024lorc,
  title={{LORC: Low-Rank Compression for LLMs KV Cache with a Progressive Compression Strategy}},
  author={Zhang, Rongzhi and Wang, Kuang and Liu, Liyuan and Wang, Shuohang and Cheng, Hao and Zhang, Chao and Shen, Yelong},
  journal={arXiv preprint arXiv:2410.03111},
  year={2024},
  url={https://arxiv.org/abs/2410.03111}
}

@inproceedings{chang2025palu,
  title={{Palu: KV-Cache Compression with Low-Rank Projection}},
  author={Chang, Chi-Chih and Lin, Wei-Cheng and Lin, Chien-Yu and Chen, Chong-Yan and Hu, Yu-Fang and Wang, Pei-Shuo and Huang, Ning-Chi and Ceze, Luis and Abdelfattah, Mohamed S and Wu, Kai-Chiang},
  booktitle={ICLR},
  volume={2025},
  pages={50222--50249},
  year={2025},
  url={https://openreview.net/forum?id=LWMS4pk2vK}
}

@article{su2024roformer,
  title={{RoFormer:: Enhanced transformer with Rotary Position Embedding}},
  author={Su, Jianlin and Ahmed, Murtadha and Lu, Yu and Pan, Shengfeng and Bo, Wen and Liu, Yunfeng},
  journal={Neurocomputing},
  volume={568},
  pages={127063},
  year={2024},
  url={https://www.sciencedirect.com/science/article/abs/pii/S0925231223011864}
}

@inproceedings{liu2024kivi,
  title={{KIVI: A Tuning-Free Asymmetric 2bit Quantization for KV Cache}},
  author={Zirui Liu and Jiayi Yuan and Hongye Jin and Shaochen Zhong and Zhaozhuo Xu and Vladimir Braverman and Beidi Chen and Xia Hu},
  booktitle={ICML},
  pages={32332--32344},
  year={2024},
  url={https://openreview.net/forum?id=L057s2Rq8O}
}

@inproceedings{xiao2023smoothquant,
  title={{SmoothQuant: Accurate and Efficient Post-Training Quantization for Large Language Models}},
  author={Xiao, Guangxuan and Lin, Ji and Seznec, Mickael and Wu, Hao and Demouth, Julien and Han, Song},
  booktitle={ICML},
  pages={38087--38099},
  year={2023},
  url={https://openreview.net/forum?id=sHfSV8eYEp}
}

@inproceedings{he2024zipcache,
  title={{ZipCache: Accurate and Efficient KV Cache Quantization with Salient Token Identification}},
  author={He, Yefei and Zhang, Luoming and Wu, Weijia and Liu, Jing and Zhou, Hong and Zhuang, Bohan},
  booktitle={NeurIPS},
  volume={37},
  pages={68287--68307},
  year={2024},
  url={https://openreview.net/forum?id=5t4ZAkPiJs}
}

@inproceedings{saxena2024eigen,
  title={{Eigen Attention: Attention in Low-Rank Space for KV Cache Compression}},
  author={Saxena, Utkarsh and Saha, Gobinda and Choudhary, Sakshi and Roy, Kaushik},
  booktitle={Findings of EMNLP},
  pages={15332--15344},
  year={2024},
  url={https://aclanthology.org/2024.findings-emnlp.899/}
}

@article{hu2025efficient_chelsea,
  title={{Efficient Long-Context LLM Inference via KV Cache Clustering}},
  author={Hu, Jie and Wang, Shengnan and He, Yutong and Gong, Ping and Yi, Jiawei and Zhang, Juncheng and Bai, Youhui and Chen, Renhai and Zhang, Gong and Li, Cheng and others},
  journal={arXiv preprint arXiv:2506.11418},
  year={2025},
  url={https://arxiv.org/abs/2506.11418}
}

@inproceedings{liu2025clusterkv,
  title={{ClusterKV: Manipulating LLM KV Cache in Semantic Space for Recallable Compression}},
  author={Liu, Guangda and Li, Chengwei and Zhao, Jieru and Zhang, Chenqi and Guo, Minyi},
  booktitle={DAC},
  pages={1--7},
  year={2025},
  url={https://ieeexplore.ieee.org/abstract/document/11132479/}
}

@inproceedings{liu2024cachegen,
  title={{CacheGen: KV Cache Compression and Streaming for Fast Large Language Model Serving}},
  author={Liu, Yuhan and Li, Hanchen and Cheng, Yihua and Ray, Siddhant and Huang, Yuyang and Zhang, Qizheng and Du, Kuntai and Yao, Jiayi and Lu, Shan and Ananthanarayanan, Ganesh and others},
  booktitle={ACM SIGCOMM},
  pages={38--56},
  year={2024},
  url={https://dl.acm.org/doi/pdf/10.1145/3651890.3672274}
}

@article{wang2024kvmerger,
  title={{Model Tells You Where to Merge: Adaptive KV Cache Merging for LLMs on Long-Context Tasks}},
  author={Wang, Zheng and Jin, Boxiao and Yu, Zhongzhi and Zhang, Minjia},
  journal={arXiv preprint arXiv:2407.08454},
  year={2024},
  url={https://arxiv.org/abs/2407.08454}
}

@inproceedings{cui2025homogeneous,
  title={{Homogeneous Keys, Heterogeneous Values: Exploiting Local KV Cache Asymmetry for Long-Context LLMs}},
  author={Cui, Wanyun and Xu, Mingwei},
  booktitle={NeurIPS},
  volume={38},
  pages={81628--81650},
  year={2025},
  url={https://openreview.net/forum?id=cvFFar8cAr}
}

@inproceedings{yuan2025weightedkv,
  title={{WeightedKV: Attention Scores Weighted Key-Value Cache Merging for Large Language Models}},
  author={Yuan, Jian and He, Ziwei and Bai, Haoli and Leng, Jingwen and Jiang, Bo},
  booktitle={ICASSP},
  pages={1--5},
  year={2025},
  url={https://ieeexplore.ieee.org/abstract/document/10889583}
}

@article{yang2025qwen2_1m,
  title={{Qwen2.5-1M Technical Report}},
  author={An Yang and Bowen Yu and Chengyuan Li and Dayiheng Liu and Fei Huang and Haoyan Huang and Jiandong Jiang and Jianhong Tu and Jianwei Zhang and Jingren Zhou and Junyang Lin and others},
  journal={arXiv preprint arXiv:2501.15383},
  year={2025},
  url={https://arxiv.org/abs/2501.15383}
}

@article{team2024qwen2,
  title={{Qwen2 Technical Report}},
  author={An Yang and Baosong Yang and Binyuan Hui and Bo Zheng and Bowen Yu and Chang Zhou and Chengpeng Li and Chengyuan Li and Dayiheng Liu and Fei Huang and Guanting Dong and others},
  journal={arXiv preprint arXiv:2407.10671},
  year={2024},
  url={https://arxiv.org/abs/2407.10671}
}

@inproceedings{wolf2020transformers,
  title={{Transformers: State-of-the-Art Natural Language Processing}},
  author={Wolf, Thomas and Debut, Lysandre and Sanh, Victor and Chaumond, Julien and Delangue, Clement and Moi, Anthony and Cistac, Pierric and Rault, Tim and Louf, R{\'e}mi and Funtowicz, Morgan and others},
  booktitle={EMNLP 2020: System Demonstrations},
  pages={38--45},
  year={2020},
  url={https://aclanthology.org/2020.emnlp-demos.6/}
}

@article{gradient_ckpt,
  title={{Training Deep Nets with Sublinear Memory Cost}},
  author={Chen, Tianqi and Xu, Bing and Zhang, Chiyuan and Guestrin, Carlos},
  journal={arXiv preprint arXiv:1604.06174},
  year={2016},
  url={https://arxiv.org/abs/1604.06174}
}

@inproceedings{penedo2024fineweb,
  title={{The FineWeb Datasets: Decanting the Web for the Finest Text Data at Scale}},
  author={Guilherme Penedo and Hynek Kydl{\'\i}{\v{c}}ek and Loubna Ben Allal and Anton Lozhkov and Margaret Mitchell and Colin Raffel and Leandro Von Werra and Thomas Wolf},
  booktitle={NeurIPS},
  volume={37},
  pages={30811--30849},
  year={2024},
  url={https://openreview.net/forum?id=n6SCkn2QaG}
}

@inproceedings{bai2024longbench,
  title={{LongBench: A Bilingual, Multitask Benchmark for Long Context Understanding}},
  author={Bai, Yushi and Lv, Xin and Zhang, Jiajie and Lyu, Hongchang and Tang, Jiankai and Huang, Zhidian and Du, Zhengxiao and Liu, Xiao and Zeng, Aohan and Hou, Lei and Dong, Yuxiao and Tang, Jie and Li, Juanzi},
  booktitle={ACL},
  pages={3119--3137},
  year={2024},
  url={https://aclanthology.org/2024.acl-long.172/}
}

@inproceedings{cai2025pyramidkv,
  title={{PyramidKV: Dynamic KV Cache Compression based on Pyramidal Information Funneling}},
  author={Zefan Cai and Yichi Zhang and Bofei Gao and Yuliang Liu and Yucheng Li and Tianyu Liu and Keming Lu and Wayne Xiong and Yue Dong and Junjie Hu and Wen Xiao},
  booktitle={COLM},
  year={2025},
  url={https://openreview.net/forum?id=ayi7qezU87}
}

@inproceedings{feng2025adakv,
  title={{Ada-KV: Optimizing KV Cache Eviction by Adaptive Budget Allocation for Efficient LLM Inference}},
  author={Yuan Feng and Junlin Lv and Yukun Cao and Xike Xie and S Kevin Zhou},
  booktitle={NeurIPS},
  volume={38},
  pages={113152--113188},
  year={2025},
  url={https://openreview.net/forum?id=tcisuhGsQZ}
}

@inproceedings{kim2025lexico,
  title={{Lexico: Extreme KV Cache Compression via Sparse Coding over Universal Dictionaries}},
  author={Junhyuck Kim and Jongho Park and Jaewoong Cho and Dimitris Papailiopoulos},
  booktitle={ICML},
  pages={30672--30687},
  year={2025},
  url={https://openreview.net/forum?id=Yh9vxlxnjA}
}

@article{liu2025deepseek32,
  title={{DeepSeek-V3.2: Pushing the Frontier of Open Large Language Models}},
  author={Liu, Aixin and Mei, Aoxue and Lin, Bangcai and Xue, Bing and Wang, Bingxuan and Xu, Bingzheng and Wu, Bochao and Zhang, Bowei and Lin, Chaofan and Dong, Chen and others},
  journal={arXiv preprint arXiv:2512.02556},
  year={2025},
  url={https://arxiv.org/abs/2512.02556}
}

@inproceedings{xiao2024infllm,
  title={{InfLLM: Training-Free Long-Context Extrapolation for LLMs with an Efficient Context Memory}},
  author={Xiao, Chaojun and Zhang, Pengle and Han, Xu and Xiao, Guangxuan and Lin, Yankai and Zhang, Zhengyan and Liu, Zhiyuan and Sun, Maosong},
  booktitle={NeurIPS},
  volume={37},
  pages={119638--119661},
  year={2024},
  url={https://openreview.net/forum?id=bTHFrqhASY}
}

@inproceedings{sun2025shadowkv,
  title={{ShadowKV: KV Cache in Shadows for High-Throughput Long-Context LLM Inference}},
  author={Hanshi Sun and Li-Wen Chang and Wenlei Bao and Size Zheng and Ningxin Zheng and Xin Liu and Harry Dong and Yuejie Chi and Beidi Chen},
  booktitle={ICML},
  pages={57355--57373},
  year={2025},
  url={https://openreview.net/forum?id=oa7MYAO6h6}
}

@inproceedings{yao2023react,
  title={{ReAct: Synergizing Reasoning and Acting in Language Models}},
  author={Yao, Shunyu and Zhao, Jeffrey and Yu, Dian and Du, Nan and Shafran, Izhak and Narasimhan, Karthik and Cao, Yuan},
  booktitle={ICLR},
  year={2023},
  url={https://iclr.cc/virtual/2023/oral/12647}
}

@inproceedings{jimenez2024swe,
  title={{SWE-bench: Can Language Models Resolve Real-world GitHub Issues?}},
  author={Jimenez, Carlos E and Yang, John and Wettig, Alexander and Yao, Shunyu and Pei, Kexin and Press, Ofir and Narasimhan, Karthik},
  booktitle={ICLR},
  volume={2024},
  pages={54107--54157},
  year={2024},
  url={https://openreview.net/forum?id=VTF8yNQM66}
}

@inproceedings{alayrac2022flamingo,
  title={{Flamingo: a Visual Language Model for Few-Shot Learning}},
  author={Alayrac, Jean-Baptiste and Donahue, Jeff and Luc, Pauline and Miech, Antoine and Barr, Iain and Hasson, Yana and Lenc, Karel and Mensch, Arthur and Millican, Katherine and Reynolds, Malcolm and others},
  booktitle={NeurIPS},
  volume={35},
  pages={23716--23736},
  year={2022},
  url={https://openreview.net/forum?id=EbMuimAbPbs}
}

@inproceedings{guha2023legalbench,
  title={{LegalBench: A Collaboratively Built Benchmark for Measuring Legal Reasoning in Large Language Models}},
  author={Guha, Neel and Nyarko, Julian and Ho, Daniel and R{\'e}, Christopher and Chilton, Adam and Chohlas-Wood, Alex and Peters, Austin and Waldon, Brandon and Rockmore, Daniel and Zambrano, Diego and others},
  booktitle={NeurIPS},
  volume={36},
  pages={44123--44279},
  year={2023},
  url={https://openreview.net/forum?id=WqSPQFxFRC}
}

@inproceedings{hao2026unix,
  title={{Uni-X: Mitigating Modality Conflict with a Two-End-Separated Architecture for Unified Multimodal Models}},
  author={Jitai Hao and Hao Liu and Xinyan Xiao and Qiang Huang and Jun Yu},
  booktitle={ICLR},
  year={2026},
  url={https://openreview.net/forum?id=IJLIYpCkwz}
}

@misc{maa_aime,
  title={{MAA Invitational Competitions}},
  author={MAA},
  howpublished={\url{https://maa.org/maa-invitational-competitions/}},
  note={Accessed: 2026-01-28. Includes the American Invitational Mathematics Examination (AIME) section.}
}

@article{qi2025deltallm,
  title={{DeltaLLM: A Training-Free Framework Exploiting Temporal Sparsity for Efficient Edge LLM Inference}},
  author={Qi, Jiawen and Gao, Chang and Ren, Zhaochun and Chen, Qinyu},
  journal={arXiv preprint arXiv:2507.19608},
  year={2025},
  url={https://arxiv.org/abs/2507.19608}
}

@inproceedings{kim2025kvzip,
  title={{KVzip: Query-Agnostic KV Cache Compression with Context Reconstruction}},
  author={Kim, Jang-Hyun and Kim, Jinuk and Kwon, Sangwoo and Lee, Jae W. and Yun, Sangdoo and Song, Hyun Oh},
  booktitle={NeurIPS},
  volume={38},
  pages={167563--167591},
  year={2025},
  url={https://openreview.net/forum?id=JFygzwx8SJ}
}

@article{yang2025qwen3,
  title={{Qwen3 Technical Report}},
  author={Yang, An and Li, Anfeng and Yang, Baosong and Zhang, Beichen and Hui, Binyuan and Zheng, Bo and Yu, Bowen and Gao, Chang and Huang, Chengen and Lv, Chenxu and others},
  journal={arXiv preprint arXiv:2505.09388},
  year={2025},
  url={https://arxiv.org/abs/2505.09388}
}

@article{bai2025qwen3,
  title={{Qwen3-VL Technical Report}},
  author={Bai, Shuai and Cai, Yuxuan and Chen, Ruizhe and Chen, Keqin and Chen, Xionghui and Cheng, Zesen and Deng, Lianghao and Ding, Wei and Gao, Chang and Ge, Chunjiang and others},
  journal={arXiv preprint arXiv:2511.21631},
  year={2025},
  url={https://arxiv.org/abs/2511.21631}
}

@article{li2024llava,
  title={{LLaVA-OneVision: Easy Visual Task Transfer}},
  author={Bo Li and Yuanhan Zhang and Dong Guo and Renrui Zhang and Feng Li and Hao Zhang and Kaichen Zhang and Peiyuan Zhang and Yanwei Li and Ziwei Liu and Chunyuan Li},
  journal={arXiv preprint arXiv:2408.03326},
  year={2024},
  url={https://arxiv.org/abs/2408.03326}
}

@inproceedings{hsieh2024ruler,
  title={{RULER}: What{\textquoteright}s the Real Context Size of Your Long-Context Language Models?},
  author={Cheng-Ping Hsieh and Simeng Sun and Samuel Kriman and Shantanu Acharya and Dima Rekesh and Fei Jia and Boris Ginsburg},
  booktitle={COLM},
  year={2024},
  url={https://openreview.net/forum?id=kIoBbc76Sy}
}

@inproceedings{chen2024image_fastv,
  title={{An Image is Worth 1/2 Tokens After Layer 2: Plug-and-Play Inference Acceleration for Large Vision-Language Models}},
  author={Chen, Liang and Zhao, Haozhe and Liu, Tianyu and Bai, Shuai and Lin, Junyang and Zhou, Chang and Chang, Baobao},
  booktitle={ECCV},
  pages={19--35},
  year={2024},
  url={https://link.springer.com/chapter/10.1007/978-3-031-73004-7_2}
}

@inproceedings{alvar2025divprune,
  title={{DivPrune: Diversity-based Visual Token Pruning for Large Multimodal Models}},
  author={Alvar, Saeed Ranjbar and Singh, Gursimran and Akbari, Mohammad and Zhang, Yong},
  booktitle={CVPR},
  pages={9392--9401},
  year={2025},
  url={https://ieeexplore.ieee.org/abstract/document/11095009}
}

@inproceedings{shen2025fastvid,
  title={{FastVID: Dynamic Density Pruning for Fast Video Large Language Models}},
  author={Leqi Shen and Guoqiang Gong and Tao He and Yifeng Zhang and Pengzhang Liu and Sicheng Zhao and Guiguang Ding},
  booktitle={NeurIPS},
  volume={38},
  pages={123553--123581},
  year={2025},
  url={https://openreview.net/forum?id=2xS4VtpApy}
}

@article{zandieh2025turboquant,
  title={{TurboQuant: Online Vector Quantization with Near-optimal Distortion Rate}},
  author={Zandieh, Amir and Daliri, Majid and Hadian, Majid and Mirrokni, Vahab},
  journal={arXiv preprint arXiv:2504.19874},
  year={2025},
  url={https://arxiv.org/abs/2504.19874}
}

@inproceedings{hooper2024kvquant,
  title={{KVQ}uant: Towards 10 Million Context Length {LLM} Inference with {KV} Cache Quantization},
  author={Coleman Richard Charles Hooper and Sehoon Kim and Hiva Mohammadzadeh and Michael W. Mahoney and Sophia Shao and Kurt Keutzer and Amir Gholami},
  booktitle={NeurIPS},
  volume={37},
  pages={1270--1303},
  year={2024},
  url={https://openreview.net/forum?id=0LXotew9Du}
}

@inproceedings{dhouib2025pact,
  title={{PACT: Pruning and Clustering-Based Token Reduction for Faster Visual Language Models}},
  author={Dhouib, Mohamed and Buscaldi, Davide and Vanier, Sonia and Shabou, Aymen},
  booktitle={CVPR},
  pages={14582--14592},
  year={2025},
  url={https://ieeexplore.ieee.org/abstract/document/11093578}
}

@article{shi2025deep,
  title={{Deep Research: A Systematic Survey}},
  author={Shi, Zhengliang and Chen, Yiqun and Li, Haitao and Sun, Weiwei and Ni, Shiyu and Lyu, Yougang and Fan, Run-Ze and Jin, Bowen and Weng, Yixuan and Zhu, Minjun and others},
  journal={arXiv preprint arXiv:2512.02038},
  year={2025},
  url={https://arxiv.org/abs/2512.02038}
}

@article{zhu2026deeprubric,
  title={{DEEPRUBRIC: Evidence-Tree Rubric Supervision for Efficient Reinforcement Learning of Deep Research Agents}},
  author={Zhu, Minghang and Wei, Chuyang and Xu, Junhao and Cheng, Yilin and Chen, Zhumin and He, Jiyan},
  journal={arXiv preprint arXiv:2606.17029},
  year={2026},
  url={https://arxiv.org/abs/2606.17029}
}

@inproceedings{
  yan2025inftythink,
  title={InftyThink: Breaking the Length Limits of Long-Context Reasoning in Large Language Models},
  author={Yuchen Yan and Yongliang Shen and Yang Liu and Jin Jiang and Mengdi Zhang and Jian Shao and Yueting Zhuang},
  booktitle={The Fourteenth International Conference on Learning Representations},
  year={2026},
  url={https://openreview.net/forum?id=T1h5em349L}
}

@inproceedings{
  yan2026inftythinkplus,
  title={InftyThink+: Effective and Efficient Infinite-Horizon Reasoning via Reinforcement Learning},
  author={Yuchen Yan and Liang Jiang and Jin Jiang and Shuaicheng Li and Zujie Wen and Zhiqiang Zhang and Jun Zhou and Jian Shao and Yueting Zhuang and Yongliang Shen},
  booktitle={Forty-third International Conference on Machine Learning},
  year={2026},
  url={https://openreview.net/forum?id=tyul8kXaJU}
}

@misc{lu2026uicopilot,
      title={UI-Copilot: Advancing Long-Horizon GUI Automation via Tool-Integrated Policy Optimization},
      author={Zhengxi Lu and Fei Tang and Guangyi Liu and Kaitao Song and Xu Tan and Jin Ma and Wenqi Zhang and Weiming Lu and Jun Xiao and Yueting Zhuang and Yongliang Shen},
      year={2026},
      eprint={2604.13822},
      archivePrefix={arXiv},
      primaryClass={cs.LG},
      url={https://arxiv.org/abs/2604.13822},
}

@inproceedings{NEURIPS2025_02d028c3,
 author = {Zhao, Haoran and Yan, Yuchen and Shen, Yongliang and Xu, Haolei and Zhang, Wenqi and Song, Kaitao and Shao, Jian and Lu, Weiming and Xiao, Jun and Zhuang, Yueting},
 booktitle = {Advances in Neural Information Processing Systems},
 editor = {D. Belgrave and C. Zhang and H. Lin and R. Pascanu and P. Koniusz and M. Ghassemi and N. Chen},
 pages = {1861--1887},
 publisher = {Curran Associates, Inc.},
 title = {Let LRMs Break Free from Overthinking via Self-Braking Tuning},
 url = {https://proceedings.neurips.cc/paper_files/paper/2025/file/02d028c3bf5d4875b5a4e4b4311087f7-Paper-Conference.pdf},
 volume = {38},
 year = {2025}
}
\bibliographystyle{IEEEtran}

\begingroup

\begin{IEEEbiography}[{\includegraphics[width=1in,height=1.25in,trim=205bp 0 205bp 0,clip,keepaspectratio]{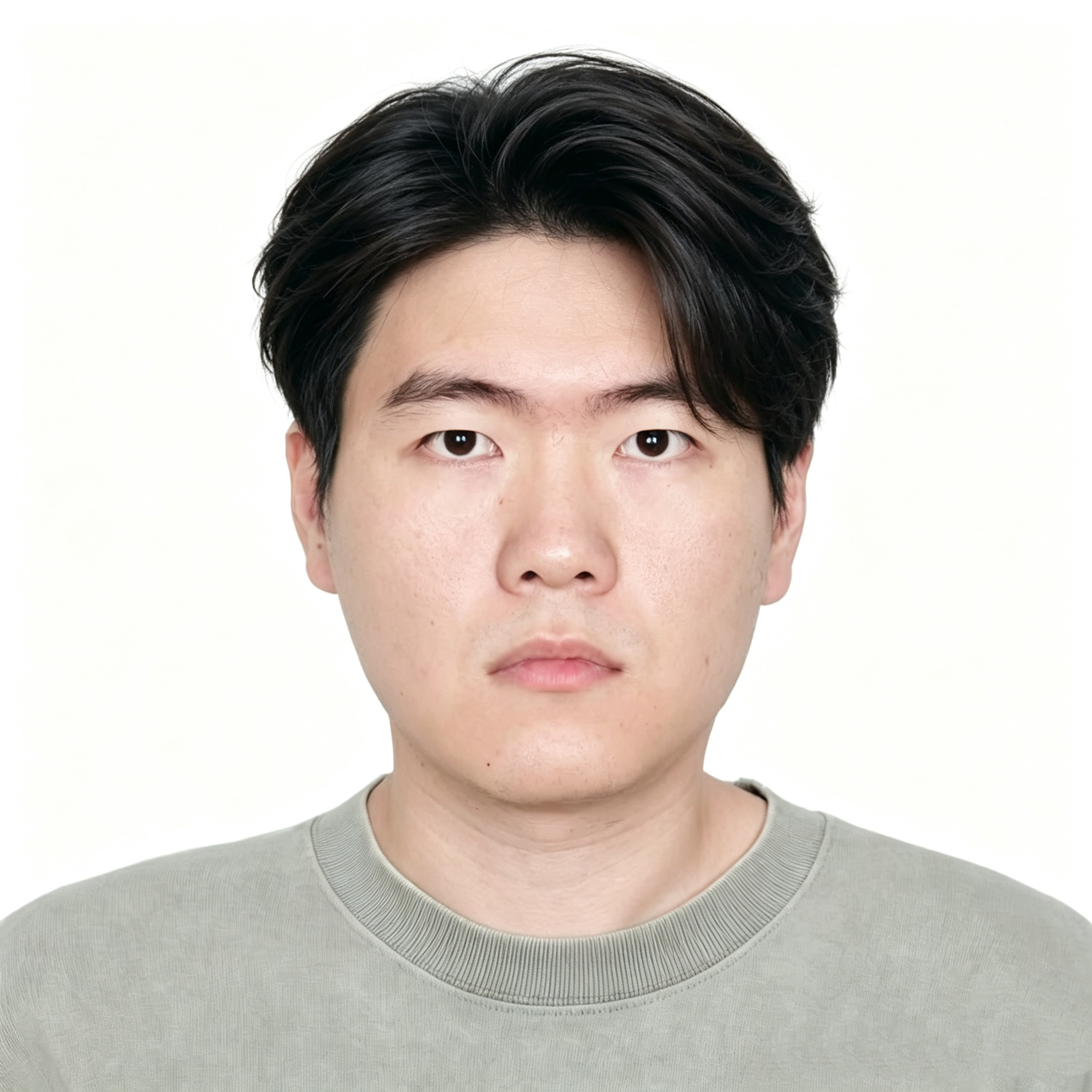}}]{Jitai Hao} received the bachelor's and master's degrees in computer science from Shandong University, Qingdao, China, in 2022 and 2025, respectively. He is currently pursuing the Ph.D. degree with the School of Intelligence Science and Engineering, Harbin Institute of Technology (Shenzhen). His research interests include efficient AI, long-context model inference, and unified multimodal models. He has published papers at leading international conferences, including ICLR, NeurIPS and ACL. His NeurIPS 2025 paper was selected as a Spotlight. During his master's studies, he received the National Scholarship. He also won two silver medals at the ACM/ICPC Asia Regional Contests.
\end{IEEEbiography}

\begin{IEEEbiography}[{\includegraphics[width=1in,height=1.25in,clip,keepaspectratio]{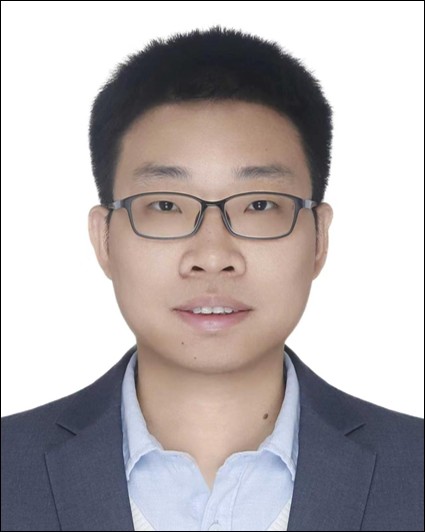}}]{Qiang Huang} received the B.Eng. and Ph.D. degrees from Sun Yat-sen University, Guangzhou, China, in 2012 and 2017, respectively. From 2017 to 2024, he was a Postdoctoral Fellow and later Senior Research Fellow at the National University of Singapore, Singapore.
He is currently a Professor and Director of the Department of Artificial Intelligence, Harbin Institute of Technology (Shenzhen), China. 
His research interests include efficient large language models, long-context inference, and multimodal learning. He has published more than 30 first- or corresponding-author papers in leading conferences and journals. 
His work received a Best Research Paper nomination at VLDB 2024, and he was recognized as an Outstanding Reviewer at ICLR 2021 and a Distinguished PC Member at ICDE 2026.
\end{IEEEbiography}

\begin{IEEEbiography}[{\includegraphics[width=1in,height=1.25in,clip,keepaspectratio]{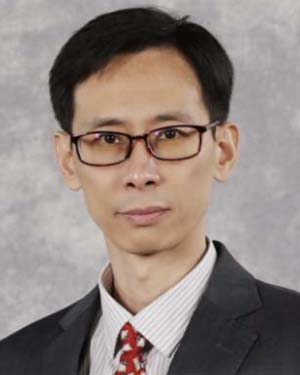}}]{Yaowei Wang} received the B.S. and M.S. degrees from Hebei University of Technology, Tianjin, China, in 1997 and 2000, respectively, and the Ph.D. degree in computer science from the University of Chinese Academy of Sciences, Beijing, China, in 2005. He is currently a Professor with the School of Computer Science and Technology, Harbin Institute of Technology (Shenzhen), Shenzhen, China. His research interests include machine learning, multimedia content analysis and understanding, and computer vision. He has authored more than 120 papers, including publications in IEEE TIP, CVPR, and ICCV. He received the Second Prize of the National Technology Invention Award in 2017 and the First Prize of the CIE Technology Invention Award in 2015. He also serves as the Chair of the IEEE Digital Retina Systems Working Group.
\end{IEEEbiography}

\begin{IEEEbiography}[{\includegraphics[width=1in,height=1.25in,trim=90bp 240bp 88bp 20bp,clip,keepaspectratio]{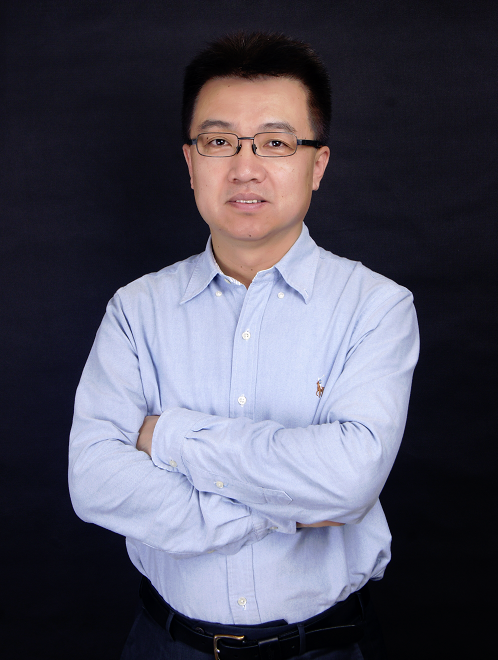}}]{Min Zhang} received the B.S., M.S., and Ph.D. degrees in computer science from Harbin Institute of Technology, Harbin, China, in 1991, 1994, and 1997, respectively. From 2003 to 2012, he was with the Institute for Infocomm Research, A*STAR, Singapore, where he headed the Statistical Machine Translation Laboratory. He joined Soochow University, Suzhou, China, as a Distinguished Professor in 2012. He is currently a Distinguished Professor and Distinguished Assistant Principal with Harbin Institute of Technology (Shenzhen), Shenzhen, China, where he serves as the Dean of the Institute of Computing and Intelligence. His research interests include machine translation, natural language processing, and large language models. He has authored more than 300 papers and received Best Paper Awards at ACM Multimedia 2022 and COLING 2022. He is an ACL Fellow and an AAIS Fellow. 
\end{IEEEbiography}

\begin{IEEEbiography}[{\includegraphics[width=1in,height=1.25in,clip,keepaspectratio]{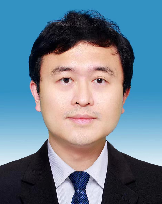}}]{Jun Yu} received the B.S. and Ph.D. degrees from Zhejiang University, Hangzhou, China, in 2003 and 2009, respectively.
From 2009 to 2011, he was a Postdoctoral Fellow at Nanyang Technological University, Singapore.
He subsequently served as an Associate Professor at Xiamen University from 2011 to 2014 and as a Professor and later Dean of the School of Computer Science at Hangzhou Dianzi University from 2014 to 2024. 
He is currently a Professor and the Dean of the School of Intelligence Science and Engineering, Harbin Institute of Technology (Shenzhen), Shenzhen, China. 
His research interests include multimodal learning, embodied intelligence, image segmentation and generation, and large language models.
Dr. Yu has authored more than 400 papers in leading journals and conferences.
He currently serves as an Associate Editor for IEEE TMM, an Editorial Board Member of Pattern Recognition and Information Sciences, and previously served as an Associate Editor for IEEE TCSVT. 
His research has received multiple Best Paper Awards, including those from IEEE TMM, IEEE TIP, and IEEE TCYB.
\end{IEEEbiography}
\vfill
\endgroup

\end{document}